\documentclass[journal]{IEEEtran}

\usepackage{cite}
\usepackage{url}
\usepackage{xcolor}
\usepackage{textcomp}

\ifCLASSINFOpdf
\usepackage[pdftex]{graphicx}
\graphicspath{{./figs/}} 
\else
\usepackage[dvips]{graphicx}
\fi
\usepackage{algorithm}
\usepackage{algorithmic}

\usepackage[caption=false,font=footnotesize,labelfont=rm,textfont=rm]{subfig}
\usepackage[colorlinks=true, linkcolor=blue, citecolor=blue, urlcolor=blue]{hyperref}
\usepackage{array}     
\usepackage{booktabs}  
\usepackage{multirow}  
\usepackage{makecell}  

\usepackage{amsmath,amssymb,amsfonts}
\usepackage{algorithmic} 
\usepackage{cuted}       

\usepackage{enumitem}
\usepackage{stfloats}    
\usepackage{ragged2e}    

\begin{document}
	
	\title{Risk-Aware Decision-Making for Autonomous Overtaking: A World Model-Based Mixture-of-Experts Framework}
	
	
    \author{Yongzhi~Liu,
            Sunan~Zhang,
            Jinchang~Xu,
            Jiawei~Wang,
            Yushu~Qiu,
            Chen~Lv,~\IEEEmembership{Senior~Member,~IEEE,}
            and~Weichao~Zhuang,~\IEEEmembership{Member,~IEEE}
    \thanks{This work was supported by the National Natural Science Foundation of China (NSFC) under Grants 52441204 and 52172383. (Corresponding author: Weichao Zhuang.)}%
    \thanks{Y. Liu, S. Zhang, J. Xu, Y. Qiu, and W. Zhuang are with the School of Mechanical Engineering, Southeast University, Nanjing 211189, China (e-mail: yongzhiliu@seu.edu.cn; wzhuang@seu.edu.cn).}%
    \thanks{J. Wang is with the Department of Civil and Environmental Engineering, University of Michigan, Ann Arbor, MI 48109 USA (e-mail: jiawe@umich.edu).}%
    \thanks{C. Lv is with the Department of Mechanical and Aerospace Engineering, Nanyang Technological University, Singapore 639798 (e-mail: lyuchen@ntu.edu.sg).}%
    }
	
	\markboth{IEEE Transactions on Intelligent Transportation Systems}%
	{Liu \MakeLowercase{\textit{et al.}}: Risk-Aware Mixture of Experts Planning via World Models}
	
	\maketitle
	
\begin{abstract}

Autonomous highway overtaking demands foresighted decision-making to handle complex interactions, stochastic traffic evolution, and temporal risk accumulation. However, standard safe reinforcement learning approaches typically rely on implicit value-based risk estimations rather than explicit dynamics modeling, thereby struggling to accurately capture complex risk propagation over multi-step horizons. This limitation frequently results in behaviors that are locally safe but induce substantial latent risks in the long term. To address this, a World Model-based Risk-aware Mixture-of-Experts (WM-RMoE) framework is proposed. First, a learned latent dynamics model facilitates parallel multi-step rollouts, elevating safety assessment from the action level to the trajectory level via cumulative risk evaluation. Second, to enhance robustness under varying interaction intensities, a hierarchical gating mechanism dynamically coordinates experts across long-horizon, short-horizon, and rule-based safety modules. Furthermore, a Gaussian Mixture Model is integrated to preserve multimodal maneuvering branches, thereby mitigating the issue of behavioral mode averaging. Experimental results demonstrate that WM-RMoE significantly outperforms representative baselines in terms of safety compliance, decision stability, and generalization capability. Furthermore, benefiting from the risk-aware formulation, the proposed framework uniquely exhibits the ability to generate foresighted and semantically distinct overtaking maneuvers across diverse traffic densities.

\end{abstract}


\section{Introduction}

\IEEEPARstart{I}{n} highway driving scenarios, autonomous vehicles must make continuous decisions—such as lane changing and merging—under stochastic traffic flows and strategic interactions \cite{Schwarting2018Planning, ShalevShwartz2016Safe}. In such tightly coupled environments, safety risks evolve dynamically throughout the interaction process rather than being determined by isolated actions, necessitating decision-making that anticipates multi-step behavioral evolution while balancing safety and efficiency \cite{Paden2016Survey, Lefevre2014Survey}.


Reinforcement learning (RL) has demonstrated potential for such sequential decision problems (e.g., TD3, PPO, SAC) \cite{Fujimoto2018TD3, Schulman2017PPO, Haarnoja2018SAC}. However, deploying purely model-free RL in safety-critical traffic remains challenging due to low sample efficiency and uncontrolled exploration \cite{Kendall2019LearningSafety, Garcia2015SafeRL}. Although Safe Reinforcement Learning (SafeRL) mitigates these issues by embedding safety constraints—via methods like Constrained Policy Optimization (CPO) \cite{Achiam2017CPO}, Lagrangian relaxation (RCPO, PID-Lagrangian) \cite{Tessler2019RCPO, Stooke2020PID}, or Lyapunov-based guarantees \cite{Chow2018Lyapunov, Zhang2026MeUAL}—most existing formulations remain predominantly action-level or expectation-based. Consequently, they often fail to adequately capture long-horizon risk accumulation and temporal risk evolution in highly interactive and uncertain settings.


Model-based reinforcement learning (MBRL) facilitates multi-step look-ahead and enhanced sample efficiency by planning within a learned dynamics model. Pioneering works like PlaNet \cite{Hafner2019PlaNet} and the Dreamer series \cite{Hafner2020Dreamer, Hafner2021DreamerV2, Hafner2023DreamerV3} have established latent space planning and long-horizon imagination as effective control paradigms. In autonomous driving, recent adaptations address domain-specific challenges: LVM \cite{Zhang2021LVM} mitigates value overestimation via a double-critic design; SEM2 \cite{Gao2024SEM2} and Iso-Dream \cite{Pan2022IsoDream} enhance robustness against visual distractors through semantic masking and dynamic decoupling, respectively; and SafeDreamer integrates Lagrangian constraints to demonstrate the viability of safety-aware world models \cite{Huang2024SafeDreamer}. Nevertheless, intrinsic model bias remains inevitable in complex traffic, and compounded rollout errors can undermine safety assessments \cite{Janner2019MBPO,Zhang2024Integration}. Critically, although the latent structure of world models computationally permits efficient batch simulation, most existing approaches restrict this capability to optimizing a single policy distribution\cite{Hafner2019PlaNet,Liu2025ITSC}. Such reliance on a unitary behavioral mode lacks the diversity required to hedge against model-reality mismatches, particularly under high-uncertainty interactions.

This reliance on unitary policies becomes particularly detrimental in strongly interactive highway scenarios, where the space of feasible decisions inherently bifurcates over long horizons. Under identical safety constraints, multiple semantically consistent yet distinct maneuver options (e.g., aggressive overtaking versus conservative following) often coexist, representing divergent interaction intents. Approximating such high-dimensional, multi-modal distributions via a unimodal parametric policy (typically a single-mode Gaussian) inevitably results in “mode averaging,” where the learned behavior regresses toward an unfeasible mean rather than selecting a contextually optimal branch \cite{Chai2019MultiPath, Codevilla2018CIL}. To resolve this structural mismatch and fully exploit the diverse rollout capabilities of world models, Mixture-of-Experts (MoE) offers a principled mechanism for conditional computation, activating specialized experts to handle distinct dynamic regimes \cite{Lepikhin2020GShard, Fedus2021Switch}. Consequently, recent end-to-end autonomous driving frameworks have begun to integrate learnable MoE routing, including dual-aware scene-adaptive routing (GEMINUS) \cite{Wan2025GEMINUS}, autoregressive trajectory planning (ARTEMIS) \cite{Feng2025ARTEMIS}, progressive expert expansion (MoPE) \cite{Cui2025MoPE}, and diffusion-based sparse routing (KDP) \cite{Xu2025KDP}.
However, existing routing policies are predominantly driven by static scene attributes or instantaneous uncertainty proxies, thereby decoupling expert selection from the long-horizon risk evolution inherent in dynamic interactions. Furthermore, if the downstream aggregation of expert behaviors lacks explicit multimodal constraints, the planning output remains susceptible to mode collapse and inadequate branch coverage. These limitations necessitate a mechanism where routing is governed by trajectory-level risk evaluation, and planning enforces distinct behavioral branches to ensure semantic consistency under safety constraints.


%
%
%

In summary, a critical methodological lacuna remains in effectively coupling trajectory-level risk filtration with robust multimodal planning. Contemporary frameworks struggle to reconcile the dual objectives of counteracting compounding latent dynamics errors and preventing the degeneration of diverse behavioral modes into a singular collapse over extended horizons. To bridge this gap, this paper presents a World Model-based Risk-aware Mixture-of-Experts (WM-RMoE) framework. By synergizing latent-space foresight with a risk-sensitive gating mechanism, the proposed architecture realizes decision-making that is strictly safety-compliant, robust against epistemic uncertainty, and adaptable to heterogeneous traffic scenarios. The primary contributions are delineated as follows:



\begin{itemize}[leftmargin=*, itemsep=1ex]
	\item[(i)] \textbf{Trajectory-level risk assessment:} A long-horizon safety filter is established via world-model rollouts to estimate cumulative risk, elevating constraint enforcement from instantaneous actions to complete trajectories to preemptively eliminate hazardous candidates.
	\item[(ii)] \textbf{Risk-driven mixture-of-experts planning:} A hierarchical planning framework is constructed that coordinates heterogeneous policies (learned, rule-based, and exploratory) via a risk-aware router, dynamically adapting expert selection to latent risk levels for balanced safety and efficiency.
	\item[(iii)] \textbf{Multimodal constrained optimization:} A Gaussian Mixture Model (GMM)-guided Cross-Entropy Method (CEM) is introduced to optimize action sequences in the latent space. By explicitly modeling the sampling distribution as a mixture of Gaussians, this approach circumvents the unimodal bias inherent in standard planners. Consequently, semantically distinct maneuver branches are effectively preserved and evaluated under strict safety constraints.
\end{itemize}


Comprehensive evaluations against representative SafeRL and MBRL baselines validate the proposed framework's superior safety compliance, task efficiency, and robustness under stochastic traffic dynamics. 

The remainder of this paper is organized as follows: Section II outlines the preliminaries; Section III details the proposed WM-RMoE framework; Sections IV and V present the experimental setup and comparative analysis; and Section VI concludes the study.

\section{PRELIMINARIES}


\subsection{CMDP Formulation for Highway Driving}
The safety-critical highway decision-making problem is formulated as a Constrained Markov Decision Process (CMDP) \cite{Altman1999CMDP}, defined by the tuple $\mathcal{M}=(\mathcal{O},\mathcal{A},P,r,c,\gamma)$.
Here, $\mathcal{O}$ and $\mathcal{A}$ denote the continuous observation and action spaces, $P$ represents the unknown transition dynamics, and $\gamma$ is the discount factor.
The reward function $r$ quantifies driving utility (e.g., efficiency and comfort), while the cost function $c$ measures safety violations such as collisions or lane departures.
The optimization objective is to find a policy $\pi(a_t|o_t)$ that maximizes the expected cumulative reward $J_r(\pi)$ subject to a discount safety cost budget $d$:
\begin{align}
	\max_{\pi} \quad & J_r(\pi) = \mathbb{E}_{\pi}\left[\sum_{t=0}^{\infty}\gamma^{t}r(o_t,a_t)\right] \\
	\text{s.t.} \quad & J_c(\pi) = \mathbb{E}_{\pi}\left[\sum_{t=0}^{\infty}\gamma_c^{t}c(o_t,a_t)\right] \le d.
\end{align}
where $\gamma_c$ reflects the temporal accumulation of safety risks.
This formulation explicitly enforces trajectory-level safety constraints within long-horizon planning tasks \cite{Ray2019SafeExploration}.

\subsection{RSSM Latent World Model}

To facilitate multi-step look-ahead prediction in dynamic traffic environments, a latent world model based on the Recurrent State-Space Model (RSSM) is employed, adhering to the training paradigm of DreamerV3 \cite{Hafner2020Dreamer,Hafner2023DreamerV3}.
The RSSM maps high-dimensional observations into compact latent state sequences to learn temporal evolution in the latent space.
Specifically, the model architecture comprises the following components parameterized by $\theta$:
\begin{align}
	\text{Sequence Model:} \quad & \hat{s}_t \sim p_{\theta}(\hat{s}_t \mid s_{t-1}, a_{t-1}) \\
	\text{Obs. Encoder:} \quad & s_t \sim q_{\theta}(s_t \mid s_{t-1}, a_{t-1}, o_t) \\
	\text{Obs. Decoder:} \quad & \hat{o}_t \sim p_{\theta}(o_t \mid s_t) \\
	\text{Reward \& Cost:} \quad & \hat{r}_t \sim p_{\theta}(r_t \mid s_t), \quad \hat{c}_t \sim p_{\theta}(c_t \mid s_t).
\end{align}
Here, the Sequence Model serves as a dynamics prior to predict the future state $\hat{s}_t$ recursively without accessing the current observation.
The Observation Encoder functions as a posterior that incorporates the real observation $o_t$ to refine the latent state estimate $s_t$.
The Observation Decoder reconstructs visual inputs to ensure semantic consistency, while the reward and cost heads predict task-relevant signals.

The model is optimized via a variational objective that maximizes the Evidence Lower Bound (ELBO), jointly minimizing reconstruction and prediction errors while regularizing the posterior toward the prior via Kullback-Leibler (KL) divergence \cite{Kingma2014VAE,Hafner2023DreamerV3}: 
\begin{equation} 
	\begin{aligned} 
		\mathcal{L}(\theta) = \mathbb{E}_{q_\theta} \Bigg[ \sum_{t} \bigg( & \ln p_\theta(o_t \mid s_t) + \ln p_\theta(r_t \mid s_t) + \ln p_\theta(c_t \mid s_t) \\
		& - \beta \, D_{\mathrm{KL}}\left(q_{\theta}(s_t \mid \cdot) \parallel p_{\theta}(s_t \mid \cdot)\right) \bigg) \Bigg]. 
	\end{aligned} 
\end{equation}


Post-training, the RSSM supports latent imagination: given $s_t$ and an action sequence $\{a_t,\ldots,a_{t+H-1}\}$, it recursively predicts $\{s_{t+1},\ldots,s_{t+H}\}$ and associated metrics $\{\hat r_{t+k},\hat c_{t+k}\}_{k=0}^{H-1}$. This mechanism serves as a differentiable and efficient backbone for the subsequent trajectory-level risk evaluation and planning.

\begin{figure*}[ht]
	\centering
	\includegraphics[width=1\textwidth]{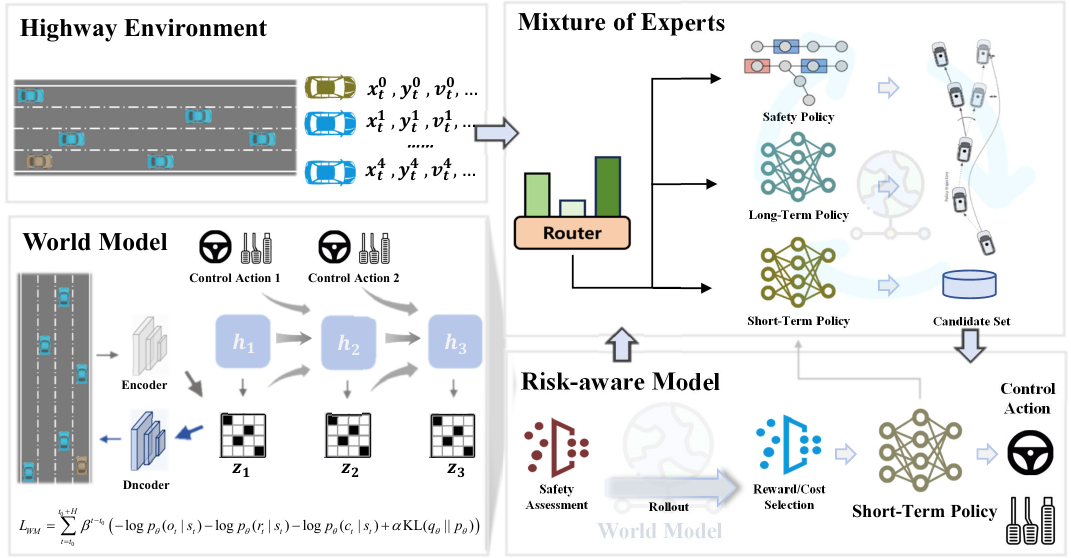}
	\caption{Schematic overview of the World Model-based Risk-aware Mixture-of-Experts (WM-RMoE) framework. 
		The architecture integrates three synergistic modules: 
		(1) Latent World Model (Left): High-dimensional traffic observations are encoded into compact latent states via a Recurrent State-Space Model (RSSM), which learns temporal dynamics ($h_t, z_t$) and optimizes variational objectives ($L_{WM}$). 
		(2) Heterogeneous Mixture of Experts (Top-Right): A risk-aware router dynamically regulates the contributions of distinct policy experts (Safety, Long-Term, and Short-Term) to construct a diverse candidate action set. 
		(3) Risk-aware Optimization (Bottom-Right): Candidate trajectories undergo multi-step parallel rollouts within the latent imagination. The optimal control action is derived through explicit safety assessment and reward/cost-based filtration, ensuring constraint satisfaction before execution.} 
	\label{fig1}
	\vspace{-10pt}
\end{figure*}

\section{METHODOLOGY}\label{AA}
\subsection{Overview of the WM-RMoE Framework} \label{sec:overview}
Building upon the CMDP formulation and the learned RSSM latent dynamics, the World Model-based Risk-aware Mixture-of-Experts (WM-RMoE) planning framework is established for highway autonomous driving. As depicted in Fig. \textcolor{blue}{\ref{fig1}}, the architecture operates through a tightly coupled pipeline of synergistic modules. 

Initially, the \textit{Latent World Model} encodes high-dimensional traffic observations into compact latent states. Conditioned on these states, the \textit{Heterogeneous Mixture-of-Experts (MoE)} module leverages a risk-aware router to blend complementary inductive biases from distinct policy experts, thereby generating a diverse set of candidate action sequences. Subsequently, within the \textit{Risk-aware Optimization} module, these candidate sequences are propagated in parallel through the latent dynamics to generate predictive rollouts. Here, a GMM-based multimodal optimizer refines the feasible solutions to distill distinct driving semantics \cite{Williams2017MPPI}. By evaluating cumulative trajectory-level rewards and costs rather than instantaneous actions, the framework enforces strict long-horizon risk-aware decision-making \cite{Sutton2018Planning, Buehler2009Planning}. Finally, the system adheres to a receding-horizon control strategy, executing only the initial action of the optimal trajectory before replanning to ensure robust adaptation to stochastic traffic evolution \cite{Mayne2000MPC}.

\subsection{Risk-Aware Mixture-of-Experts in Latent Planning Space}
\subsubsection{Design of Heterogeneous Experts}

To reconcile traffic efficiency, local robustness, and safety fallback under stochastic highway dynamics, three complementary policy components are instantiated within the latent planning space: a long-term learned expert, a short-term memory expert, and a rule-based safety expert.
Each expert is formalized as a mapping from the latent state $s_t$ to a stochastic or deterministic action distribution, denoted as $\pi_i(\cdot \mid s_t)$.
These experts are queried in parallel within the unified planning framework to generate candidate proposals, thereby incorporating heterogeneous inductive biases across varying time scales and risk preferences.
This parallel architecture enables the planner to synthesize diverse behavioral priors at each decision step, significantly enhancing adaptability and safety margins in complex interactive scenarios.


\paragraph{Long-term learned expert}
The long-term learned expert, denoted as $\pi_{\mathrm{L}}(a \mid s; \phi)$, is designated to pursue long-horizon return maximization within the CMDP formulation.
It aims to optimize the cumulative task performance over an infinite horizon while adhering to expected safety cost constraints.
This expert is implemented using a Lagrangian-variant of the Soft Actor-Critic (SAC-Lag) architecture, comprising a stochastic actor, dual critics for reward $Q^{r}_{\mathrm{L}}$ and cost $Q^{c}_{\mathrm{L}}$, and a learnable Lagrange multiplier $\lambda$.
The multiplier $\lambda$ dynamically modulates the trade-off between task performance and constraint violation based on the degree of safety infringement.
Consequently, $\pi_{\mathrm{L}}$ serves as the primary performance-driven engine during planning, providing goal-directed and efficient action proposals that exploit the learned latent dynamics.


\paragraph{Short-Term Memory Expert}
To preserve maneuver continuity and reduce computational overhead, the short-term memory expert exploits the temporal consistency of driving behaviors.
Rather than initiating the action search from scratch, this expert retrieves the optimized multimodal solution space from the immediately preceding decision step ($t-1$).
Specifically, the feasible action distribution is modeled via a Gaussian Mixture Model (GMM) capturing $M$ distinct semantic modes:
\begin{equation}
	p_{t-1}(\mathbf{a})=\sum_{m=1}^{M}\pi_m\,\mathcal{N}(\mathbf{a}\mid \mu_m,\Sigma_m),
\end{equation}
where $\mathbf{a}$ denotes the continuous action vector, while $\pi_m$, $\mu_m$, and $\Sigma_m$ represent the respective mixing weight, mean vector, and covariance matrix of the $m$-th mode.
Acting as an instantaneous warm-start prior for the current state $s_t$, this distribution is refined through minor stochastic perturbations.
This memory-driven propagation explicitly transfers historical driving semantics, ensuring robust adaptation to interactive uncertainties without demanding deep predictive rollouts.


\paragraph{Rule-Based Safety Expert}\label{Safe_benchmark}
The rule-based safety expert serves as an interpretable fallback, ensuring kinematic feasibility independent of the learned latent representation.
Unlike the learning-based components, this expert derives deterministic actions directly from the raw traffic observations $o_t$ using the Intelligent Driver Model (IDM) \cite{Treiber2000IDM} for longitudinal control and the MOBIL criterion \cite{Kesting2007MOBIL} for lateral decisions.
To integrate these deterministic outputs into the unified probabilistic planning interface, the rule-based action is mathematically formulated as the mean of a tightly bounded Gaussian distribution:
\begin{equation}
	\pi_{\mathrm{R}}(a_t \mid o_t) = \mathcal{N}\!\left(a_t \mid g_{\mathrm{IDM\text{-}MOBIL}}(o_t), \Sigma_{\epsilon}\right),
\end{equation}
where $g_{\mathrm{IDM\text{-}MOBIL}}(\cdot)$ denotes the mapping from observations to reference kinematic commands, and $\Sigma_{\epsilon}$ represents a minimal structural covariance.
Rather than introducing erratic exploration noise, this design maintains representational consistency across all heterogeneous experts. It strictly bounds the sampling space, enabling the trajectory optimizer to evaluate valid, rule-compliant local variations within the latent imagination space without compromising the baseline safety guarantee.


Collectively, these three heterogeneous experts offer complementary strategic advantages: the \emph{long-term learned policy} pursues global value maximization through latent foresight; the \emph{short-term memory module} preserves maneuver continuity via temporal solution propagation; and the \emph{rule-based fallback} ensures deterministic kinematic safety. These distinct behavioral profiles are subsequently fused by a dynamic, risk-aware gating mechanism.

\begin{figure}[ht]
	\centering
	\includegraphics[width=0.4\textwidth]{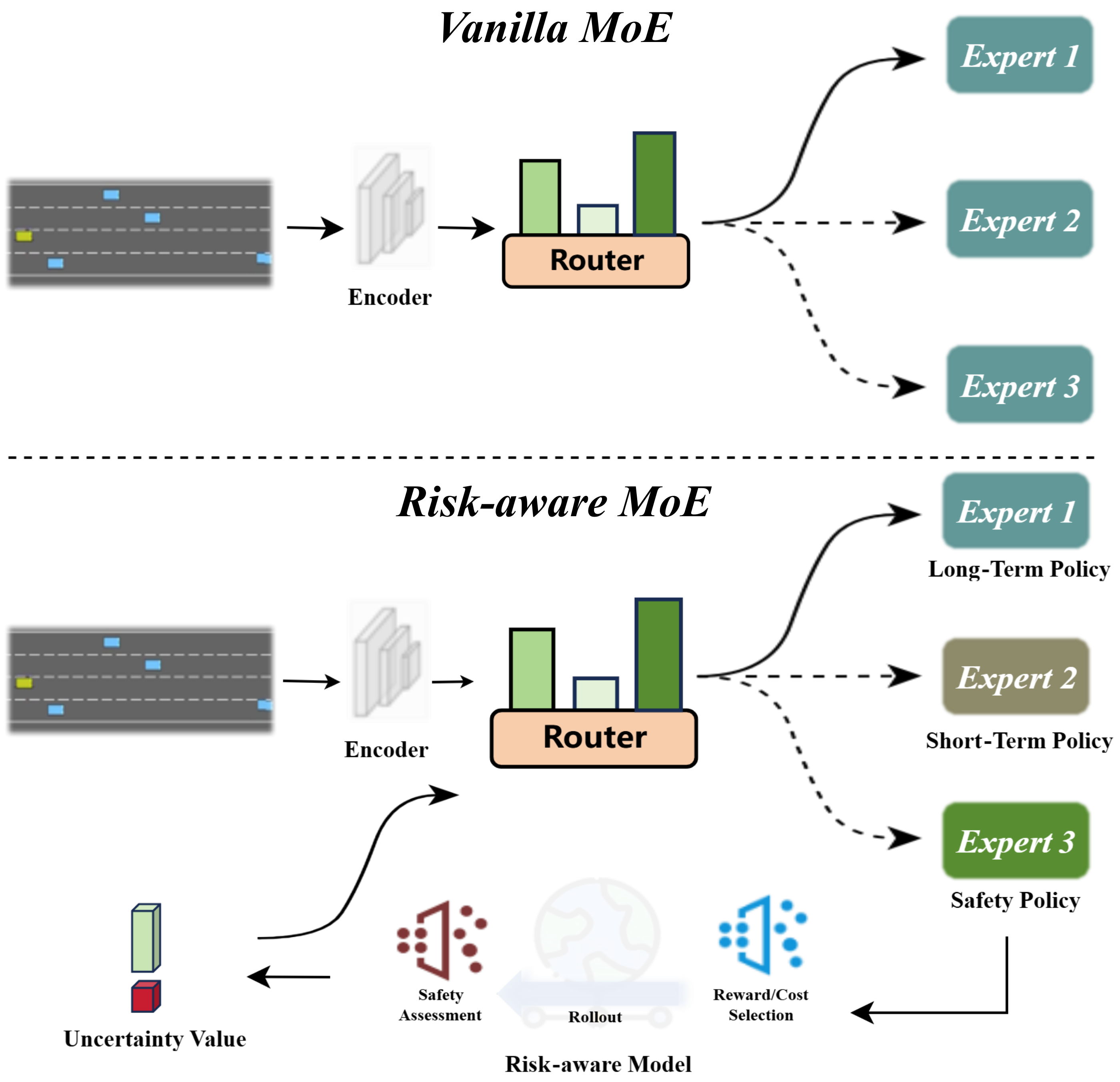}
	\caption{Comparison between vanilla MoE and the proposed risk-aware MoE.} 
	\label{moe_arch} 
	\vspace{-10pt} 
\end{figure}

\subsubsection{Risk-Aware Expert Gating with Uncertainty Feedback}


To orchestrate candidate generation under stochastic traffic evolution, a learnable risk-aware gating network $g_{\psi}$ is instantiated. As illustrated in Fig. \ref{moe_arch}, this mechanism distinguishes itself from standard feed-forward routers by incorporating a closed-loop uncertainty feedback. It acts as a soft scheduler, dynamically allocating the computational sampling budget among heterogeneous experts based on both the current environmental context and the reliability of the previous planning iteration.


Formally, the gating network conditions on the current latent state $s_t$ and the epistemic uncertainty metric $\mathcal{U}_{t-1}$, which is derived from the multimodal trajectory optimization process at the preceding step. Specifically, the metric $\mathcal{U}_{t-1}$ quantifies the spatial dispersion of the feasible elite trajectories. A broad dispersion implies a lack of consensus among the predictive optimal plans, which typically arises from highly unpredictable surrounding traffic interactions. Consequently, this epistemic entropy serves as a robust surrogate for evaluating planning risk. Furthermore, to preserve temporal consistency, the preceding distribution parameters $G_{t-1}$ are additionally incorporated into the gating input. The routing operation is formulated as:
\begin{equation}
	w_t = g_{\psi}(s_t, G_{t-1}, \mathcal{U}_{t-1}) \in \Delta^{K-1},
\end{equation}
where $\Delta^{K-1} = \{w \in \mathbb{R}^K \mid w_i \ge 0, \sum_{i=1}^K w_i = 1\}$ denotes the $(K-1)$-dimensional probability simplex, ensuring that the dynamically allocated expert weights are strictly non-negative and mathematically sum to one. The uncertainty feedback $\mathcal{U}_{t-1}$ is critical: a high uncertainty value drives the router to shift probability mass $w_t$ towards the deterministic rule-based safety expert, whereas low uncertainty encourages exploitation via the learned expert. Consequently, the final proposal distribution becomes a risk-modulated ensemble:
\begin{equation}
	q(a_t \mid s_t) = \sum_{i=1}^{K} w_{t,i} \cdot \pi_i(a_t \mid \mathcal{X}_{t,i}),
	\label{eq:mixture_dist}
\end{equation}
where $\mathcal{X}_{t,i}$ denotes the specific input context for the $i$-th expert.
Specifically, to decouple kinematic safety from latent representation errors, the input domain is defined as heterogeneous: $\mathcal{X}_{t,i} = s_t$ for the long-term learned and short-term memory experts, relying on the world model's state inference; whereas $\mathcal{X}_{t,i} = o_t$ for the rule-based safety expert, which operates directly on raw observations.

This design ensures that the system adaptively contracts its exploration space in high-risk scenarios while expanding it during safe navigation.

\begin{figure*}[ht]
	\centering
	\includegraphics[width=1\textwidth]{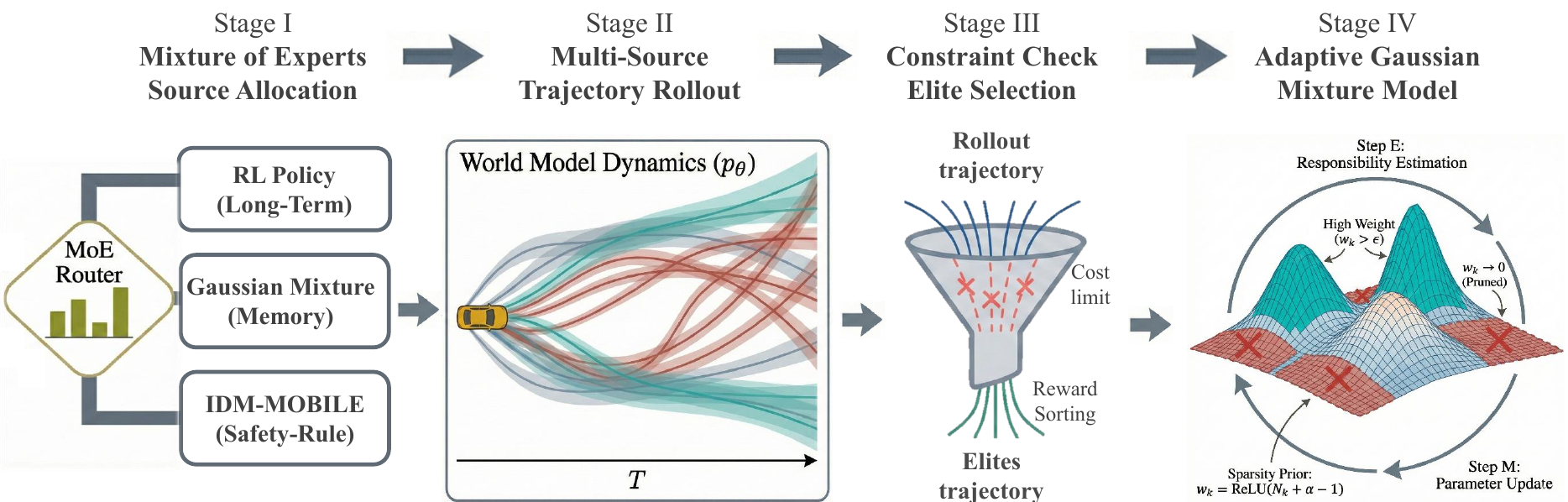}
	\caption{Overview of the proposed GMM-guided constrained cross-entropy planning (GMM-CCEM) in latent space. 
		At each decision step, candidate action sequences are sampled and rolled out in the learned world model to obtain predicted reward--cost sequences. 
		Feasible trajectories are filtered by a trajectory-level risk budget, and the remaining elites are refit with a Gaussian mixture model to update the sampling distribution, preserving multi-branch behaviors and mitigating mode averaging.}
	\label{fig3}
	\vspace{-10pt}
\end{figure*}


\subsection{Multimodal Risk-Constrained Trajectory Optimization}

Building upon the heterogeneous action candidates allocated by the MoE module, the framework transitions into the multimodal trajectory optimization phase. As illustrated in Fig. \ref{fig3}, this specific optimization process is structured into three sequential internal steps: Multi-Source Rollout, Constraint Check, and Adaptive GMM Refinement. These components formulate a closed-loop iteration designed to maximize long-term rewards while strictly adhering to safety constraints.

\subsubsection{Stage II: Multi-Source Trajectory Rollout}
In this stage, $N$ candidate action sequences $\{a^{(n)}\}_{n=1}^N$ are sampled from the hybrid proposal distribution $q(a_t \mid s_t)$ constructed by the experts. These candidates are propagated in parallel through the learned world model dynamics:
\[
s_{t+k+1}^{(n)} \sim p_\theta(s_{t+k+1} \mid s_{t+k}^{(n)}, a_{t+k}^{(n)}),
\]
yielding predicted trajectories of rewards $\hat{r}$ and costs $\hat{c}$. This step transforms the latent intent into explicit future state evolutions, allowing the planner to evaluate the long-horizon consequences of the mixed expert strategies.

\subsubsection{Stage III: Constraint Check and Elite Selection}
To enforce safety strictly, a two-step trajectory-level filtration and sorting mechanism is executed. First, candidate action sequences are rolled out in the latent space up to a maximum horizon $H$. Modulated by the termination prediction head of the world model, the effective accumulation of metrics mathematically truncates upon predicting a terminal state (e.g., a collision), thereby accommodating variable-length valid trajectories within parallel batch processing. A candidate is deemed \emph{feasible} only if its expected cumulative discounted cost satisfies a predefined safety budget $d_H$:
\begin{equation}
	\mathcal{F} = \left\{ a^{(n)} \;\middle|\; \mathbb{E} \left[ \sum_{k=0}^{H-1} \gamma_c^k \hat{c}_{t+k}^{(n)} \right] \le d_H \right\},
\end{equation}
where $\gamma_c \in (0,1]$ denotes the cost discount factor. Trajectories violating this constraint are systematically pruned. Second, the remaining feasible candidates within the set $\mathcal{F}$ are sorted based on their expected cumulative discounted return, defined as $\hat{R}^{(n)} = \mathbb{E} \left[ \sum_{k=0}^{H-1} \gamma^k \hat{r}_{t+k}^{(n)} \right]$. The top-ranking subset, representing predictive maneuvers that are both strictly safe and high-performing, constitutes the elite set $\mathcal{E}$.

\subsubsection{Stage IV: Adaptive Gaussian Mixture Modeling}
Given that the action candidates originating from the heterogeneous experts are inherently generated via Gaussian-parameterized policy heads, the aggregated elite set $\mathcal{E}$ naturally conforms to a mixture of Gaussian distributions. To accurately capture the multimodal nature of these elite behaviors (e.g., distinct semantic branches such as lane-changing versus car-following), $\mathcal{E}$ is explicitly fitted with a Gaussian Mixture Model (GMM) rather than a standard unimodal distribution. This structural alignment ensures mathematical consistency during density estimation. The parameters of the updated distribution—comprising mixture weights $\pi_m$, means $\mu_m$, and covariances $\Sigma_m$—are subsequently estimated via the Expectation-Maximization (EM) algorithm:
\begin{equation}
	q^*(a) = \sum_{m=1}^M \pi_m \mathcal{N}(a \mid \mu_m, \Sigma_m).
\end{equation}
This multimodal refinement prevents mode collapse and preserves diverse feasible strategies.
Crucially, to close the loop with the risk-aware gating network (Stage I), the planning uncertainty $\mathcal{U}_t$ is quantified as the entropy of this final distribution:
\begin{equation}
	\mathcal{U}_t = \mathbb{H}[q^*] \approx -\sum_{m=1}^M \pi_m \ln \left( \sum_{j=1}^M \pi_j e^{-D_{KL}(\mathcal{N}_m || \mathcal{N}_j)} \right).
\end{equation}
This metric serves as a retrospective feedback signal: high entropy implies a dispersed solution space with potential conflicts, prompting the router to contract the exploration space in the subsequent timestep.

\subsection{Optimization Objectives and Training Procedure}
\label{sec:learning}

\subsubsection{Analytic Lagrangian Policy Optimization in Latent Space}

To solve the CMDP formulation defined in Sec. II-A, the constrained optimization problem is transformed into an unconstrained dual problem using the Lagrangian relaxation method.
Leveraging the differentiable latent dynamics of the world model, the long-term learned expert $\pi_{\mathrm{L}}$ is optimized via analytic gradient propagation, ensuring sample-efficient convergence to the safety-constrained optimal policy.

Two critic networks, the reward value $v_{\psi}^r(s)$ and the safety cost value $v_{\psi}^c(s)$, are trained to estimate the expected objectives $J_r$ and $J_c$ from the current state.
To balance bias and variance, the regression targets are constructed using $\lambda$-returns computed over finite-horizon imagined trajectories $\{\hat{s}_t, \hat{r}_t, \hat{c}_t\}_{t=k}^{k+H}$:
\begin{equation}
	\begin{split}
		\mathcal{L}_{\mathrm{critic}}(\psi) = \frac{1}{2}  \mathbb{E}_{\tau_{\mathrm{imag}}} \Big[ & \left( v_{\psi}^r(\hat{s}_k) - V_k^r \right)^2 \\
		& + \left( v_{\psi}^c(\hat{s}_k) - V_k^c \right)^2 \Big],
	\end{split}
\end{equation}
where $V_k^{\cdot}$ denotes the computed $\lambda$-return target.
This estimation provides the actor with global performance and risk signals necessary for long-horizon planning.

The actor $\pi_{\phi}$ and the Lagrange multiplier $\lambda$ constitute a primal-dual game.
The actor (primal) minimizes the Lagrangian objective, while the multiplier (dual) maximizes the penalty for constraint violations.
Crucially, gradients are backpropagated through the differentiable transition function $p_\theta(\hat{s}_{t+1}|\hat{s}_t, \hat{a}_t)$ to the policy parameters $\phi$:
\begin{equation}
	\begin{split}
		\mathcal{L}_{\mathrm{actor}}(\phi) = - \mathbb{E}_{\tau_{\mathrm{imag}}} \sum_{t=k}^{k+H} \gamma^{t-k} \Big( & v_{\psi}^r(\hat{s}_t) - \text{sg}(\lambda) v_{\psi}^c(\hat{s}_t) \\
		& + \eta \mathcal{H}(\pi_\phi) \Big).
	\end{split}
\end{equation}
Simultaneously, the Lagrange multiplier $\lambda$ is updated via dual gradient ascent to enforce the safety budget $d$:
\begin{equation}
	\mathcal{L}_{\mathrm{lag}}(\lambda) = \lambda \Big( d - \mathbb{E}\big[ v_{\psi}^c(\hat{s}_k) \big] \Big).
\end{equation}
Here, $\text{sg}(\cdot)$ denotes the stop-gradient operator to stabilize the adversarial training.
This analytic optimization ensures that the learned expert strictly adheres to the trajectory-level safety constraints imposed by the highway driving task.

\subsection{Risk-Guided Distillation of the Gating Network}

While the GMM-CCEM planning (Sec. III-C) generates high-quality, risk-constrained decisions, its iterative optimization is computationally intensive for real-time inference. To bridge the gap between planning precision and control frequency, the gating network $g_{\psi}$ is trained via knowledge distillation to mimic the expert selection logic of the planner.
However, treating the uncertainty metric $\mathcal{U}_t$ merely as a state input often leads to gradient attenuation, failing to capture the causal necessity of safety fallback in uncertain scenarios.
To address this, a **risk-guided target rectification** mechanism is introduced to explicitly bias the supervision signal.

The baseline supervision target, $\mathbf{y}_t^{raw}$, is derived from the empirical statistics of the elite set $\mathcal{E}_t$ generated in the planning phase:
\begin{equation}
	y_{t, k}^{raw} = \frac{1}{|\mathcal{E}_t|} \sum_{\tau^{(n)} \in \mathcal{E}_t} \mathbb{I}(k^{(n)} = k),
\end{equation}
where $\mathbb{I}(\cdot)$ is the indicator function and $k^{(n)}$ denotes the expert index of the $n$-th elite trajectory.
To actively enforce safety awareness, this target is rectified using the planning uncertainty $\mathcal{U}_t$. Let $k_{\mathrm{safe}}$ denote the index of the rule-based safety expert. The rectified target $\mathbf{y}_t^*$ is formulated as a risk-modulated mixture:
\begin{equation}
	\mathbf{y}_t^* = (1 - \sigma(\mathcal{U}_t)) \cdot \mathbf{y}_t^{raw} + \sigma(\mathcal{U}_t) \cdot \mathbf{1}_{k_{\mathrm{safe}}},
	\label{eq:target_rectification}
\end{equation}
where $\mathbf{1}_{k_{\mathrm{safe}}}$ is a one-hot vector peaking at the safety expert, and $\sigma(\cdot)$ is a modulation function that saturates to 1 when $\mathcal{U}_t$ exceeds a risk threshold.
This mechanism acts as a robust inductive bias, guaranteeing that the router learns to prioritize the safety expert in highly uncertain scenarios, independent of stochastic fluctuations in elite selection.

Finally, to prioritize learning in critical driving scenarios (e.g., cut-ins or merge conflicts), an advantage-weighted cross-entropy objective is employed. An advantage term $A_t$ scales the gradient updates based on the relative value of the planned trajectory. The total distillation loss is defined as:
\begin{equation}
	\mathcal{L}_{\mathrm{gate}}(\psi) = \mathbb{E}_{\mathcal{D}} \left[ - A_t \sum_{k=1}^K y_{t, k}^* \log w_{t, k} - \beta \mathcal{H}(w_t) \right],
\end{equation}
where $w_t$ is the network output and $\mathcal{H}(w_t)$ serves as entropy regularization to prevent premature collapse. Through this objective, the risk-aware planning logic is effectively distilled into the efficient feed-forward network.

\begin{figure}[ht]
	\centering
	\includegraphics[width=0.48\textwidth]{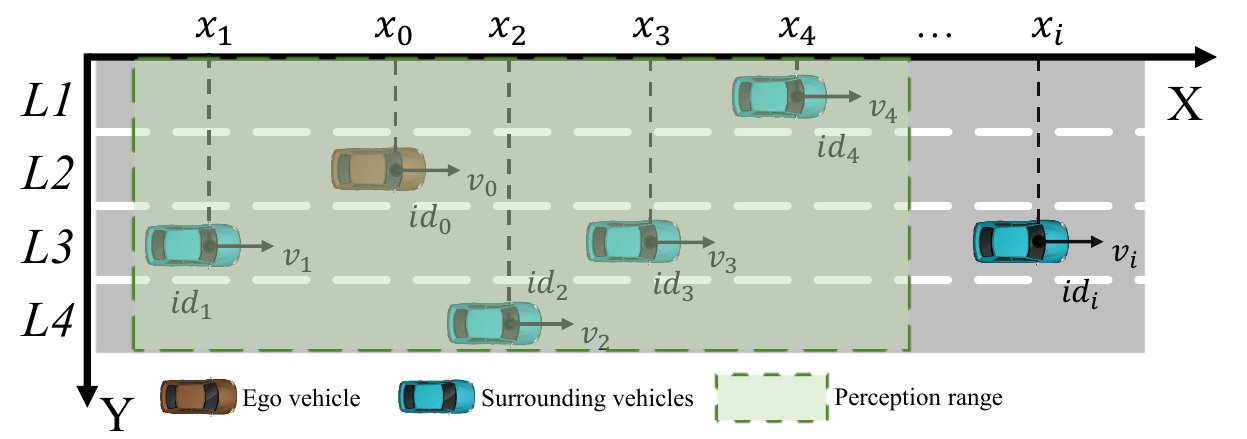}
	\caption{Schematic illustration of the four-lane highway driving scenario. The designated perception range defines the local observability boundary, within which the ego vehicle accurately acquires the full kinematic states (e.g., position, velocity, and heading) of surrounding traffic entities to construct the current observation.}
	\label{fig4}
	\vspace{-10pt}
\end{figure}

\section{EXPERIMENTAL SETUP} \label{sec:exp}
The experimental validation is conducted on the open-source highway-env platform \cite{Leurent2018Highway} configured as a four-lane unidirectional highway to emulate high-speed driving dynamics. As depicted in Fig. \ref{fig4}, the Ego Vehicle (EV) is required to execute lane-keeping and overtaking maneuvers while interacting with stochastic surrounding traffic. To rigorously model the trade-off between efficiency and safety, the task is formulated as a Constrained Markov Decision Process (CMDP), optimizing the policy against both reward maximization and cost constraints.


\subsection{State and Action Space Definitions} \label{sec:state_action}
The state space $\mathcal{S}$ is constructed assuming fully observable kinematics of the EV and surrounding vehicles. The comprehensive state vector $S$ concatenates the EV state $s_e$ and the relative states of the $K$ nearest neighbors $\{s_i\}_{i=1}^K$ within the perception range $[L_{min}, L_{max}]$:
\begin{equation}
	S = [s_e, s_1, \dots, s_K]^\top,
\end{equation}
where $K=4$ is the maximum number of tracked neighbors. The specific components are defined as:
\begin{equation}
	\begin{cases}
		s_e = [x_e, y_e, v_{x_e}, v_{y_e}, \cos\phi_e, \sin\phi_e] \\
		s_i = [\Delta x_{i}, \Delta y_{i}, \Delta v_{x_{i}}, \Delta v_{y_{i}}, \cos\phi_i, \sin\phi_i]
	\end{cases}
\end{equation}
Here, $(x, y)$ and $(v_x, v_y)$ denote positions and velocities, respectively, and $\phi$ represents the heading angle. The term $\Delta(\cdot)_i$ signifies the relative value of the $i$-th neighbor with respect to the EV.
To handle variable traffic density, zero-padding is applied when fewer than $K$ vehicles are detected. All state features are normalized to $[-1, 1]$ to facilitate training stability.

The action space $\mathcal{A}$ governs the continuous longitudinal acceleration $a$ and steering angle $\delta$:
\begin{equation}
	A = [a, \delta]^\top.
\end{equation}
Policy outputs are squashed to $[-1, 1]$ via a $\tanh$ activation function and subsequently mapped to physically feasible kinematic bounds for motion planning execution.

\subsection{Reward and Cost Function Formulation}
To induce efficient and compliant driving behaviors while adhering to safety constraints, the optimization objective is composed of a dense reward function and a composite cost signal.

The total reward $r_t$ aggregates driving efficiency and passenger comfort:
\begin{equation}
	r_t = w_e r_{\mathrm{effi}} + w_c r_{\mathrm{comf}}.
	\label{eq:reward_total}
\end{equation}
The efficiency term $r_{\mathrm{effi}}$ incentivizes high-speed cruising by normalizing the ego velocity $v_e$ relative to the task limits $[v_{\min}, v_{\max}]$:
\begin{equation}
	r_{\mathrm{effi}} = \frac{2(v_e - v_{\min})}{v_{\max} - v_{\min}} - 1.
	\label{eq:reward_effi}
\end{equation}
Simultaneously, passenger comfort $r_{\mathrm{comf}}$ is quantified by penalizing the magnitude of control variations, specifically the jerk $|\Delta a|$ and steering rate $|\Delta \delta_f|$:
\begin{equation}
	r_{\mathrm{comf}} = - \left( w_a |\Delta a| + w_s |\Delta \delta_f| \right).
	\label{eq:reward_comfort}
\end{equation}

Within the SafeRL framework, a safety cost $c_t$ is imposed to penalize hazardous states. The total cost is defined as:
\begin{equation}
	c_t = c_{\mathrm{DRF}} + c_{\mathrm{out}} + c_{\mathrm{col}}.
	\label{eq:cost_total}
\end{equation}
Where, $c_{\mathrm{DRF}}$ represents the Dynamic Risk Field (DRF) \cite{Li2022Continuous}, which continuously evaluates environmental risks. It integrates static risks from road boundaries and dynamic risks from surrounding traffic using Gaussian distributions based on relative kinematics. Complementing this continuous metric, $c_{\mathrm{out}}$ and $c_{\mathrm{col}}$ impose discrete penalties (set to 5.0) upon the activation of lane violations and collisions, respectively.
Consequently, the agent is trained to maximize the expected cumulative reward subject to the long-term cost constraint $d$.

\subsection{Implementation and PIL Validation Platform} \label{sec:implementation}

The computational experiments are executed on a workstation equipped with an Intel i7-14700 CPU, 32 GB RAM, and an NVIDIA RTX 4060 GPU. The simulation environment is built upon the open-source \textit{highway-env} platform. Episodes are configured with a duration of 20 seconds and a control frequency of 10 Hz. To emulate realistic human-like driving dynamics, surrounding vehicles are governed by the Intelligent Driver Model (IDM)~\cite{Treiber2000IDM} and the MOBIL criterion~\cite{Kesting2007MOBIL}, which regulate longitudinal car-following and lateral lane-change behaviors, respectively. Furthermore, to introduce authentic traffic stochasticity, the behavioral parameters of these models (e.g., target velocity, safe time headway, and politeness factor) are randomly sampled from predefined empirical distributions during each episode initialization. Quantitative evaluation is performed over 100 stochastic trials using metrics including success rate, average velocity, and cumulative cost.

\begin{figure}[ht]
	\centering
	\includegraphics[width=0.48\textwidth]{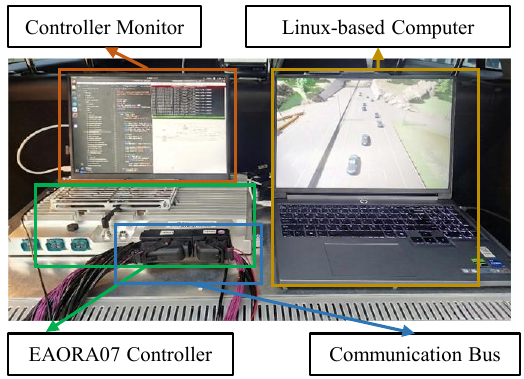}
	\caption{Processor-in-the-Loop (PIL) testing setup for real-time policy evaluation.}
	\label{PIL-fig}
	\vspace{-5pt}
\end{figure}

To validate the real-time feasibility of the proposed framework, a Processor-in-the-Loop (PIL) testing platform is established, as illustrated in Fig.~\ref{PIL-fig}. The pre-trained policy is deployed on an EAORA07 edge computing controller, which integrates dual NVIDIA Orin NX chips for onboard inference. A Linux-based host computer running the \textit{highway-env} simulation serves as the environment node. Data exchange between the edge controller (the agent) and the host (the environment) is facilitated via Ethernet using the Real-Time Publish-Subscribe (RTPS) protocol, ensuring the low-latency synchronization of state observations and control commands.

\subsection{Baselines and Evaluation Metrics} \label{sec:baselines_metrics}

To ensure a fair and comprehensive evaluation of the proposed framework, comparative experiments are conducted against a diverse set of baselines, ranging from rule-based controllers to state-of-the-art reinforcement learning algorithms.

\vspace{0.3em}
\noindent \textbf{1) Comparative Baselines:}
The selected evaluation targets encompass three distinct methodological paradigms—rule-based, model-free, and model-based approaches—to ensure a comprehensive comparative analysis.
\begin{itemize}
	\item \textbf{IDM-MOBIL}: The deterministic kinematic model defined in Sec. \ref{Safe_benchmark}, serving as the foundational rule-based safety benchmark.
	\item \textbf{SAC}: Soft Actor-Critic, a standard off-policy algorithm that maximizes entropy and expected return without explicit safety constraints.
	\item \textbf{SAC-Lag}: A constrained variant of SAC incorporating Lagrangian relaxation to enforce cost limits during the learning phase.
	\item \textbf{CVPO}: Constrained Variational Policy Optimization, a robust trust-region method designed for safe reinforcement learning.
	\item \textbf{DreamerV3}: A leading model-based architecture leveraging latent world models for long-horizon prediction~\cite{Hafner2023DreamerV3}.
	\item \textbf{SafeDreamer}: A safety-aware extension of DreamerV3 that strictly integrates cost constraints into the latent planning process~\cite{Huang2024SafeDreamer}.
\end{itemize}

\vspace{0.2em}
\noindent \textbf{2) Evaluation Metrics:}
Quantitative assessment is conducted through five principal metrics, holistically characterizing driving efficiency, algorithmic robustness, and strict safety adherence.
\begin{itemize}
	\item \textit{Average Episode Reward (AER)} and \textit{Average Speed (AS)}: Measure the overall task efficiency and the capability to maintain optimal traffic flow.
	\item \textit{Average Travel Distance (ATD)}: Quantifies the mean longitudinal distance traversed by the ego vehicle within a single evaluation episode.
	\item \textit{Success Rate (SR)}: Defined as the ratio of completely collision-free episodes, providing a direct and uncompromising safety assessment.
	\item \textit{Average Episode Cost (AEC)}: Quantifies the magnitude of constraint violations, mathematically reflecting the adherence to predefined safety boundaries.
\end{itemize}

\begin{table*}[t]
	\centering
	\caption{Quantitative Performance Comparison Across Different Traffic Densities}
	\label{tab:density_comparison}
	\resizebox{\textwidth}{!}{%
		\setlength{\tabcolsep}{5pt}
		\begin{tabular}{lccccccccc}
			\toprule
			\multirow{2}{*}{\textbf{Algorithm}} & \multicolumn{3}{c}{\textbf{Low Density}} & \multicolumn{3}{c}{\textbf{Medium Density}} & \multicolumn{3}{c}{\textbf{High Density}} \\
			\cmidrule(lr){2-4} \cmidrule(lr){5-7} \cmidrule(lr){8-10}
			& \textbf{SR (\%)} $\uparrow$ & \textbf{AER} $\uparrow$ & \textbf{AEC} $\downarrow$ & \textbf{SR (\%)} $\uparrow$ & \textbf{AER} $\uparrow$ & \textbf{AEC} $\downarrow$ & \textbf{SR (\%)} $\uparrow$ & \textbf{AER} $\uparrow$ & \textbf{AEC} $\downarrow$ \\
			\midrule
			\textbf{WM-RMoE (Ours)} & \textbf{100.0} $\pm$ 0.0 & 175.4 $\pm$ 2.1 & \textbf{8.2} $\pm$ 0.4 & 97.8 $\pm$ 2.2 & 166.8 $\pm$ 1.5 & \textbf{11.3} $\pm$ 0.6 & 91.6 $\pm$ 3.5 & \textbf{152.3} $\pm$ 3.1 & \textbf{17.8} $\pm$ 1.5 \\
			SafeDreamer & 91.2 $\pm$ 1.5 & 155.6 $\pm$ 2.4 & 10.5 $\pm$ 0.8 & 85.0 $\pm$ 1.6 & 135.7 $\pm$ 1.3 & 14.9 $\pm$ 0.6 & 78.5 $\pm$ 4.2 & 128.4 $\pm$ 4.5 & 21.6 $\pm$ 2.1 \\
			DreamerV3 & 90.5 $\pm$ 2.1 & 162.3 $\pm$ 3.5 & 12.4 $\pm$ 1.8 & 83.0 $\pm$ 4.2 & 151.8 $\pm$ 2.9 & 24.3 $\pm$ 2.3 & 65.2 $\pm$ 5.5 & 115.6 $\pm$ 6.2 & 23.9 $\pm$ 4.8 \\
			SAC-Lag & 86.1 $\pm$ 3.5 & 170.5 $\pm$ 4.2 & 14.8 $\pm$ 1.2 & 82.0 $\pm$ 5.1 & 165.1 $\pm$ 5.2 & 17.8 $\pm$ 2.8 & 62.4 $\pm$ 8.5 & 135.2 $\pm$ 7.1 & 25.4 $\pm$ 3.6 \\
			CVPO & 68.4 $\pm$ 4.0 & 165.2 $\pm$ 4.8 & 14.2 $\pm$ 1.1 & 55.0 $\pm$ 6.0 & 132.3 $\pm$ 8.9 & 16.3 $\pm$ 1.7 & 55.8 $\pm$ 10.2 & 120.4 $\pm$ 9.5 & 18.5 $\pm$ 3.2 \\
			SAC & 89.0 $\pm$ 3.0 & \textbf{178.5} $\pm$ 4.5 & 15.6 $\pm$ 1.5 & 81.0 $\pm$ 4.3 & \textbf{167.3} $\pm$ 4.8 & 22.1 $\pm$ 1.0 & 58.0 $\pm$ 9.5 & 125.8 $\pm$ 7.8 & 20.2 $\pm$ 5.5 \\
			IDM-MOBIL & \textbf{100.0} $\pm$ 0.0 & --- & --- & \textbf{100.0} $\pm$ 0.0 & --- & --- & \textbf{100.0} $\pm$ 0.0 & --- & --- \\
			\bottomrule
		\end{tabular}%
	}
\end{table*}

\section{EXPERIMENTAL RESULTS AND ANALYSIS}

\subsection{Comparative Analysis of Learning Performance}
\begin{figure}[ht]
	\centering
	\includegraphics[width=0.495\textwidth]{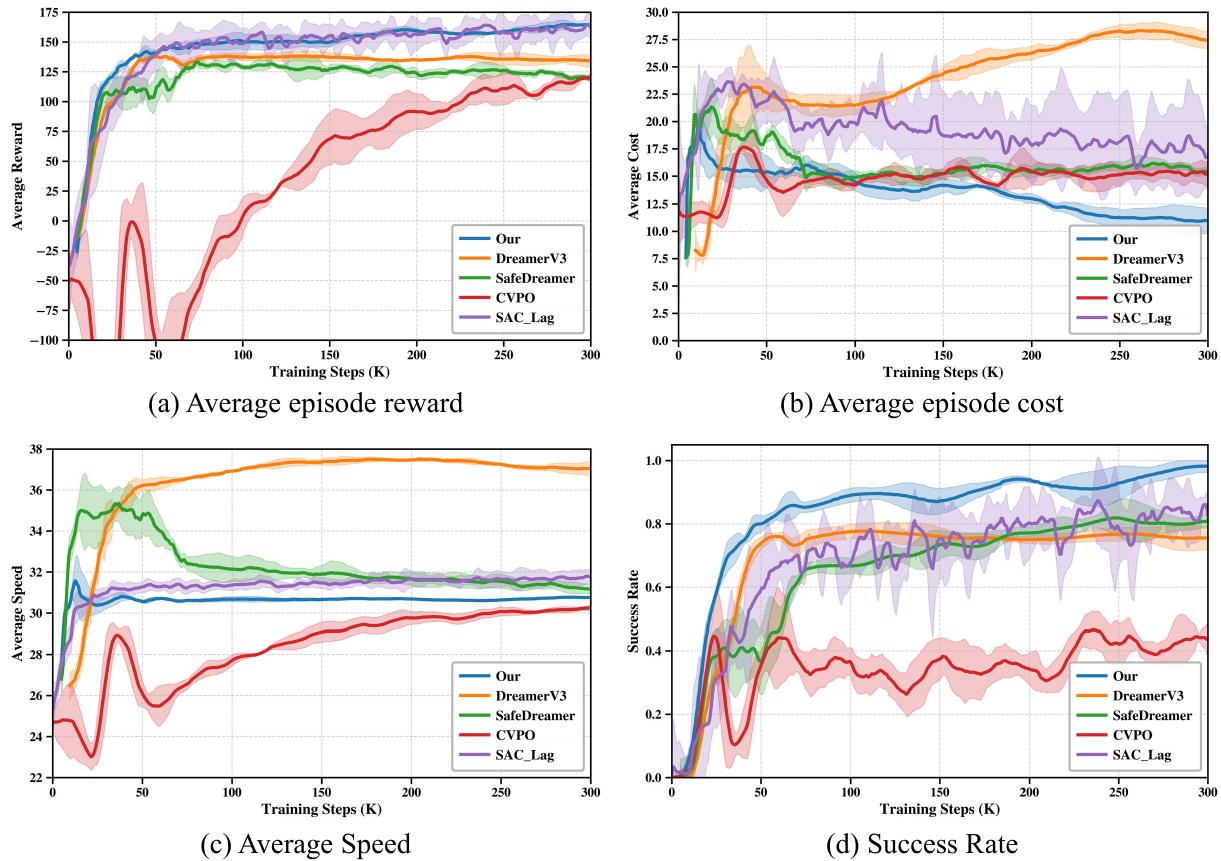}
	\caption{Comparative learning curves of WM-RMoE and baseline algorithms over 300k training steps. Shaded regions indicate standard deviation across 5 seeds.}
	\label{baselines}
	\vspace{-12pt}
\end{figure}

To evaluate the sample efficiency and convergence stability of the proposed method, all algorithms were independently trained over $3 \times 10^5$ steps with five random seeds. The comparative learning curves are presented in Fig. \ref{baselines}.
WM-RMoE demonstrates superior convergence characteristics across all metrics. Specifically, in terms of Average Episode Reward, the proposed method achieves the fastest ascent and stabilizes at the highest asymptotic value, indicating effective policy optimization. In contrast, model-free baselines (SAC, CVPO) exhibit slower learning rates and larger variances.
Regarding safety, as reflected by the \textit{Average Cost} and \textit{Success Rate}, WM-RMoE rapidly converges to a minimal cost level while maintaining a near-perfect success rate. Baseline methods, particularly the unconstrained DreamerV3, maintain high costs throughout training, suggesting a failure to effectively internalize safety constraints.
Notably, while DreamerV3 maintains a high \textit{Average Speed}, it comes at the expense of safety, whereas WM-RMoE balances efficiency and safety by stabilizing within a reasonable high-speed range.
This performance advantage is attributed to the risk-aware gating mechanism, which dynamically shifts authority to the safety expert during critical learning phases, preventing catastrophic exploration and facilitating stable gradient updates.

\subsection{Robustness Evaluation under Varying Traffic Densities}
To rigorously assess the robustness and adaptability of the proposed method against varying traffic complexities, post-training evaluations were conducted across three explicitly quantified density levels. These levels are parameterized by the spatial concentration of surrounding vehicles relative to the nominal training density $\rho_0$ (e.g., measured in vehicles per kilometer): low ($\rho = 0.5\rho_0$), medium ($\rho = \rho_0 \pm 10\%$), and high ($\rho = 1.5\rho_0$). Table \ref{tab:density_comparison} summarizes the statistical outcomes from 100 independent trials per condition. During the inference phase, the WM-RMoE framework consistently executes online receding-horizon planning, actively leveraging the learned world model for latent trajectory imagination and dynamic expert routing.

\begin{figure}[ht]
	\centering
	\includegraphics[width=0.495\textwidth]{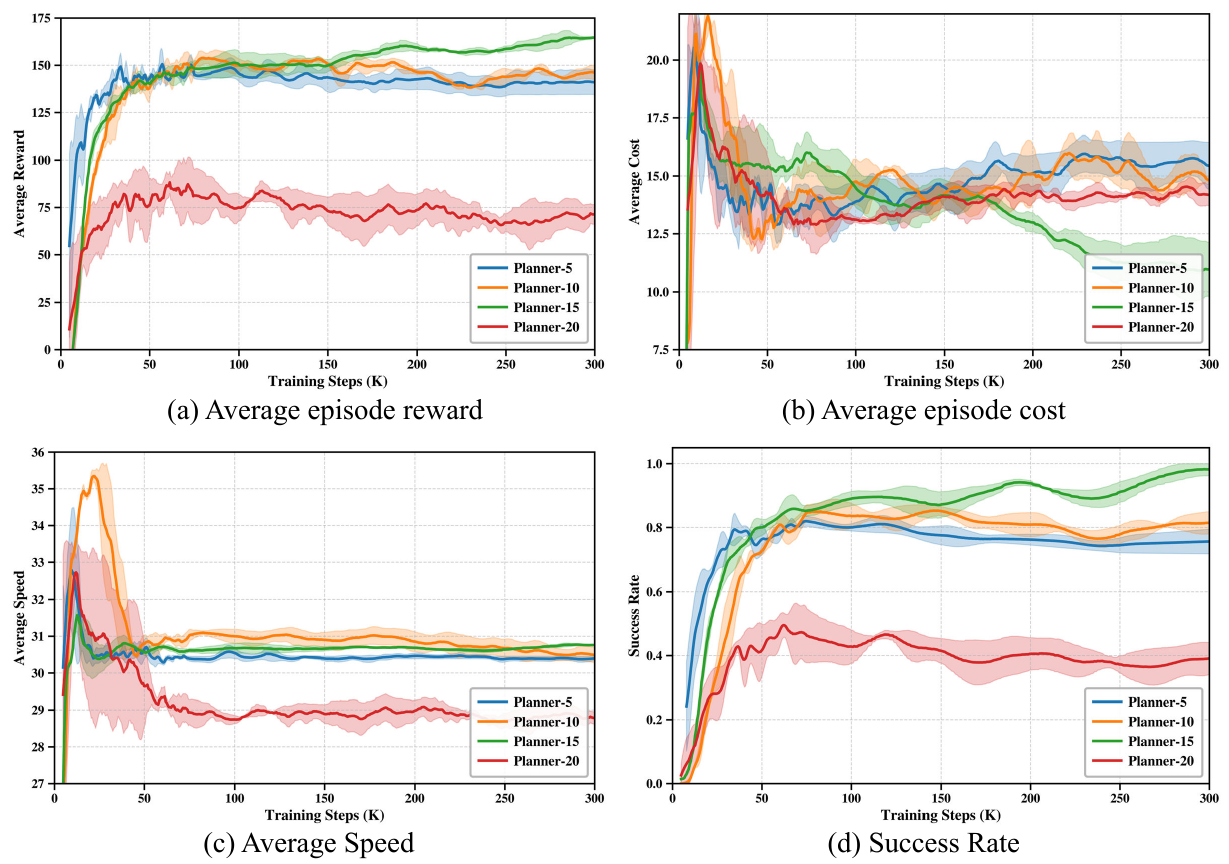}
	\caption{Learning dynamics under different planning horizons $H \in \{5,10,15,20\}$. $H=15$ offers the optimal trade-off between stability and long-term foresight.}
	\label{horizon}
	\vspace{-13pt}
\end{figure}

In low-density scenarios, most algorithms achieve high success rates; however, WM-RMoE uniquely combines absolute safety (100\% SR) with superior efficiency, outperforming the conservative SafeDreamer baseline. As traffic density intensifies to medium and high levels, the performance of the comparative methods degrades significantly due to highly coupled vehicle interactions and compounded uncertainties. Conversely, WM-RMoE exhibits robust adaptability. Even in the highly congested regime ($\rho = 1.5\rho_0$), it maintains a 91.6\% success rate—substantially surpassing the second-best SafeDreamer (78.5\%)—while simultaneously achieving the highest average reward (152.3) and the lowest cumulative cost (17.8). This structural robustness empirically validates the efficacy of the online MoE architecture, wherein the risk-aware router effectively prunes hazardous trajectory branches in complex interactive environments, thereby ensuring an optimal and strict trade-off between driving efficiency and physical safety.

\subsection{Ablation Analysis of Key Mechanisms}

To systematically dissect the contribution of each component within the WM-RMoE framework, ablation studies are conducted focusing on the planning horizon and the mixture-of-experts architecture.

\textbf{Impact of Planning Horizon.}
The planning horizon $H$ determines the foresight capability of the latent rollout.
As illustrated in Fig. \ref{horizon}, comparative experiments with $H \in \{5, 10, 15, 20\}$ reveal a distinct trade-off between foresight and error accumulation.
Short horizons ($H=5, 10$) yield rapid initial gains but plateau early, as the agent fails to anticipate long-term risks such as distant congestion. Conversely, an overly long horizon ($H=20$) introduces compounding prediction errors in the latent space, leading to reward oscillation and instability.
The configuration $H=15$ achieves the optimal balance, demonstrating steady convergence to the highest reward and success rate (near 1.0). This indicates that a moderate horizon sufficiently captures traffic dynamics while mitigating the variance of long-term imagination.

\begin{table}[t]
	\centering
	\caption{Ablation Study on MoE Components}
	\label{tab:ablation_compact}
	\resizebox{\linewidth}{!}{%
		\setlength{\tabcolsep}{4pt}
		\begin{tabular}{lccccc}
			\toprule
			\textbf{Variants} & \textbf{AER} $\uparrow$ & \textbf{AEC} $\downarrow$ & \textbf{SR (\%)} $\uparrow$ & \textbf{AS (m/s)} $\uparrow$ & \textbf{ATD (m)} $\uparrow$ \\
			\midrule
			w/o Router & 155.4 $\pm$ 5.2 & 14.5 $\pm$ 1.8 & 89.2 $\pm$ 3.5 & 29.1 $\pm$ 1.2 & 545.6 $\pm$ 15.4 \\
			w/o RL Expert & 148.6 $\pm$ 8.4 & 17.2 $\pm$ 3.1 & 88.1 $\pm$ 4.2 & 30.1 $\pm$ 1.8 & 520.1 $\pm$ 20.4 \\
			w/o GMM Expert & 125.5 $\pm$ 12.6 & 22.4 $\pm$ 4.5 & 74.5 $\pm$ 6.7 & 31.5 $\pm$ 2.9 & 420.5 $\pm$ 45.2 \\
			w/o Rule Expert & 142.1 $\pm$ 3.1 & 21.1 $\pm$ 0.9 & 85.4 $\pm$ 2.8 & \textbf{32.3} $\pm$ 1.1 & 460.2 $\pm$ 12.5 \\
			\midrule
			\textbf{WM-RMoE} & \textbf{166.8} $\pm$ 1.5 & \textbf{11.3} $\pm$ 0.6 & \textbf{97.8} $\pm$ 2.2 & 30.8 $\pm$ 0.1 & \textbf{602.4} $\pm$ 2.7 \\
			\bottomrule
		\end{tabular}%
	}
\end{table}

\begin{figure}[ht]
	\centering
	\includegraphics[width=0.45\textwidth]{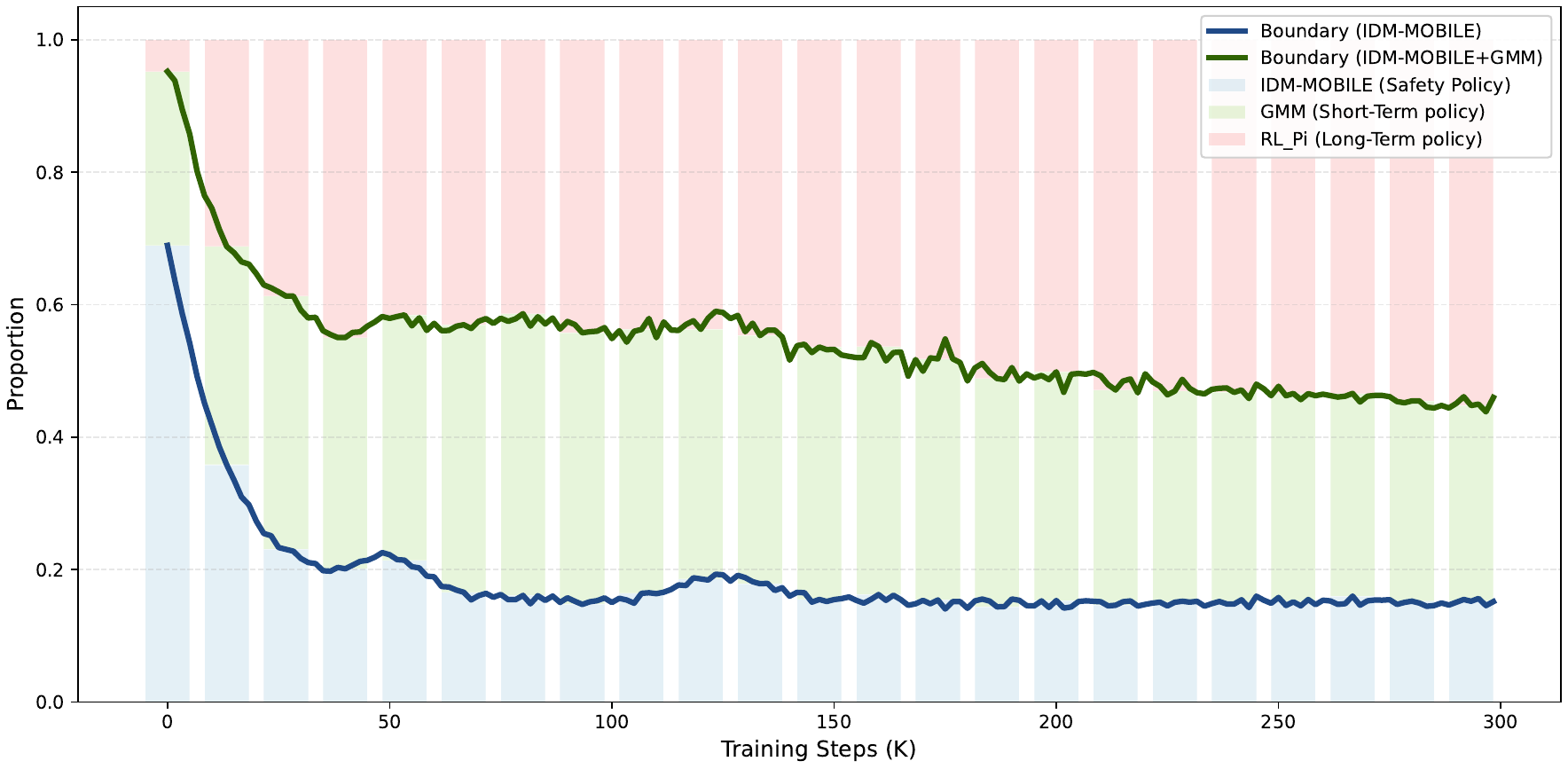}
	\caption{Evolution of routing weights during training. The system transitions from rule-dominated safety priors to RL-dominated efficiency optimization.}
	\label{MoE_Episode}
	\vspace{-5pt}
\end{figure}

\begin{figure*}[ht]
	\centering
	\includegraphics[width=0.95\textwidth]{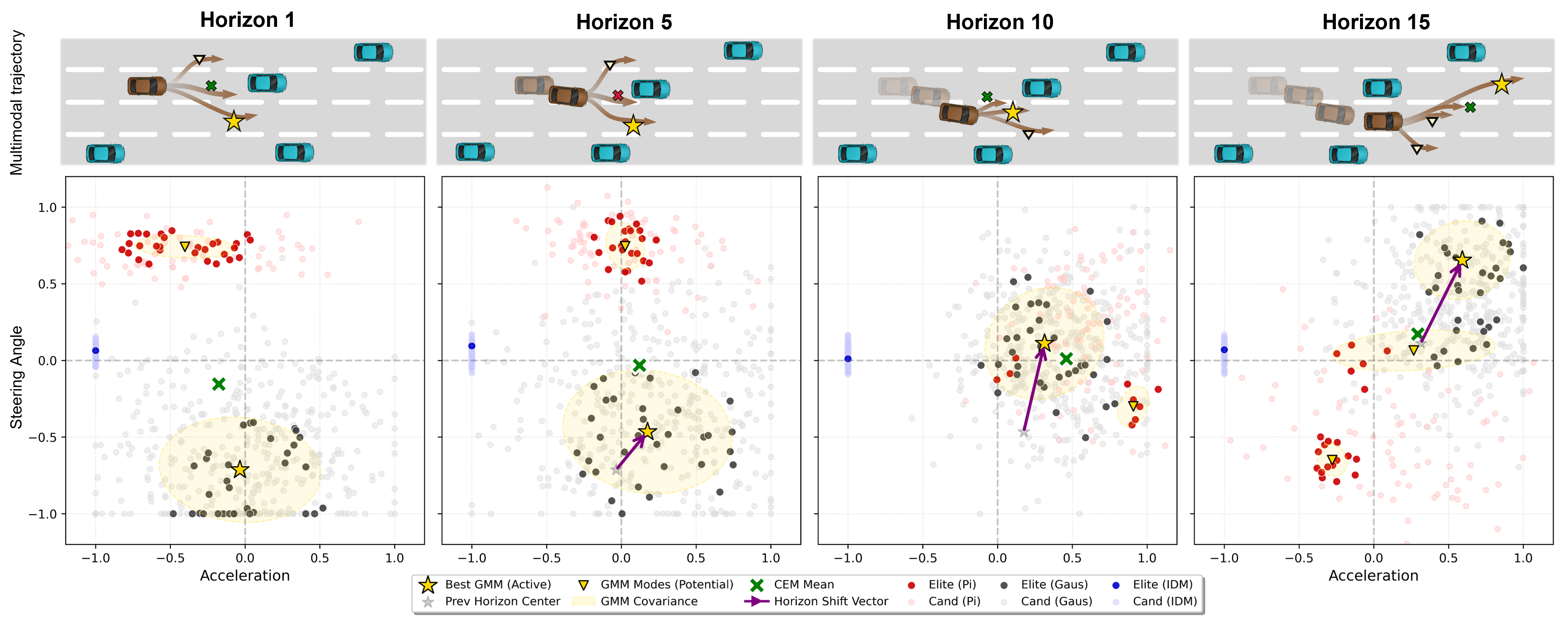}
	\caption{Dynamic evolution of multimodal decision-making across sequential environmental time steps ($t \in \{1, 15, 30, 45\}$). Top: Projected latent trajectory branches during an overtaking maneuver. Bottom: Corresponding real-time distributions of parallel-sampled action candidates in the steering-acceleration plane. Translucent and solid points denote initial priors and elite subsets, respectively. The green ``X'' illustrates hazardous mode collapse in standard CEM due to the arithmetic averaging of distinct choices. Conversely, covariance ellipses demonstrate GMM-CCEM's preservation of multimodal branches, with purple vectors tracking the continuous temporal shift of the optimal mode center (yellow star) to ensure kinematically safe transitions.}
	\label{WM_horizon}
	\vspace{-10pt}
\end{figure*}

\textbf{Contribution of Heterogeneous Experts.}
The architectural efficacy of the MoE strategy is systematically validated by ablating individual experts and the dynamic routing mechanism, with quantitative results summarized in Table \ref{tab:ablation_compact}. 
Specifically, ablating the rule-based expert (\textit{w/o Rule Expert}) induces overly aggressive maneuvers; while achieving the highest average speed (32.3 m/s), it precipitates a severe deterioration in safety constraints (cost surging to 21.1), substantiating its function as an indispensable kinematic safety bound. 
Excluding the memory-based GMM expert (\textit{w/o GMM Expert}) yields the lowest success rate (74.5\%) and heightened algorithmic instability. Without this component to actively preserve multimodal action hypotheses, the planner becomes highly susceptible to hazardous mode collapse, frequently converging toward suboptimal or kinematically infeasible solutions. 

\begin{figure}[ht]
	\centering
	\includegraphics[width=0.49\textwidth]{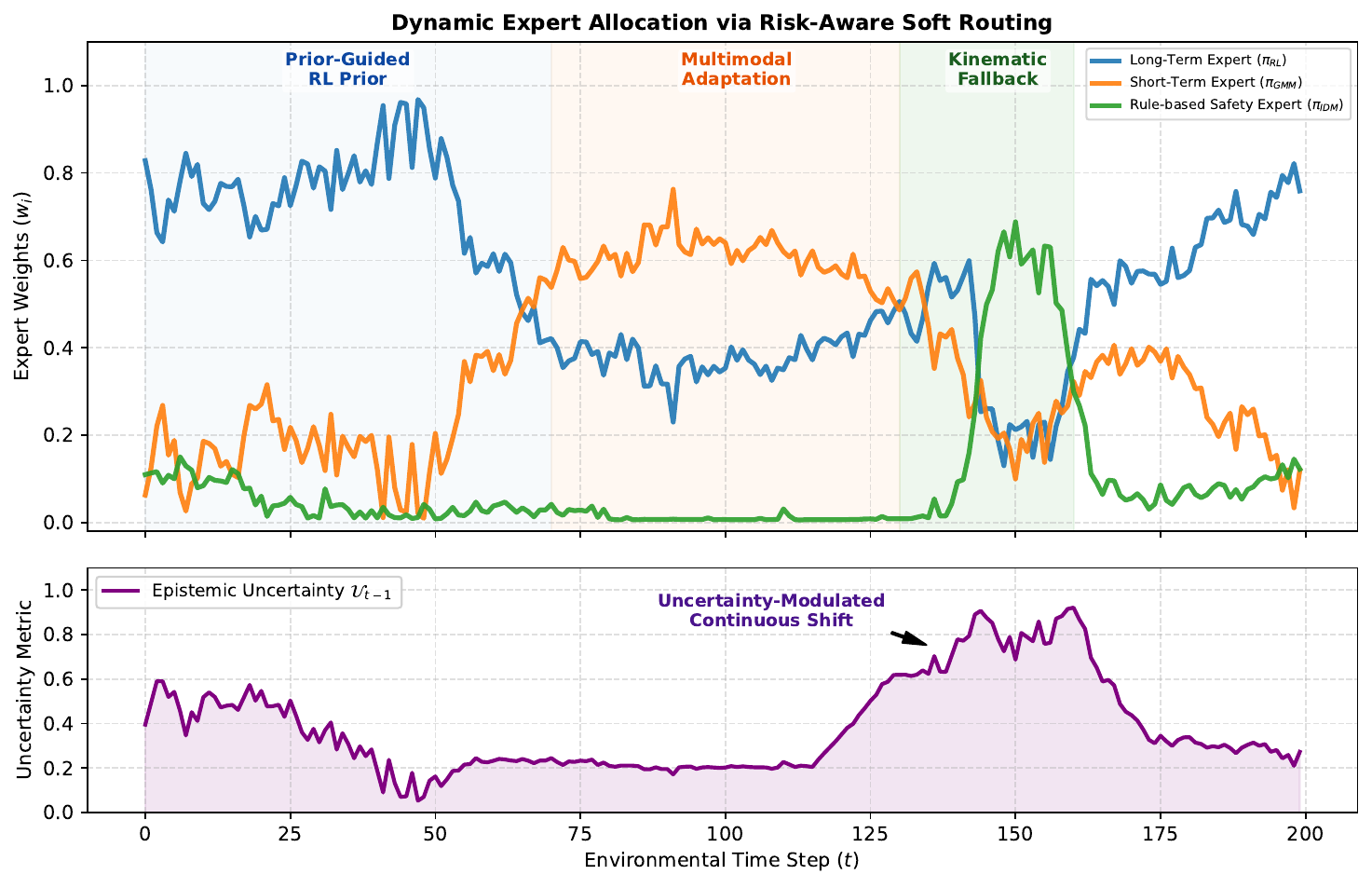}
	\caption{Risk-aware MoE routing in a test episode: expert weights (top) and epistemic uncertainty with a handover threshold (bottom). Exceeding the threshold increases the safety-expert weight.}
	\label{MoE_step} 
	\vspace{-10pt} 
\end{figure}

Furthermore, the absence of the long-term RL expert (\textit{w/o RL Expert}) results in overly conservative behavioral patterns that, while maintaining baseline safety, fail to optimize task efficiency (AER dropping to 148.6). 
Finally, replacing the dynamic router with a static uniform weight distribution (\textit{w/o Router}) uniformly degrades all performance metrics, unequivocally validating the critical necessity of the context-aware, state-dependent expert scheduling mechanism.

The evolution of expert utilization is further analyzed in Fig. \ref{MoE_Episode}.
In the early training phase, the rule-based and GMM experts dominate, providing a safe behavioral prior that accelerates convergence.
As training progresses, the router adaptively shifts authority to the RL expert (increasing weight), reflecting a transition from imitation of rules to the exploitation of learned value functions.
This dynamic scheduling significantly enhances robustness in complex traffic scenarios.


\subsection{Qualitative Analysis of Adaptive Decision-Making}


\paragraph{Intra-Episode Dynamic Expert Allocation}
To elucidate the dynamic scheduling mechanism, Fig. \ref{MoE_step} depicts the temporal evolution of expert weights during a representative lane-changing episode. In the initial phase ($t=0\sim60$), the long-term learned policy dominates (blue curve), providing a global value foundation for rapid cruising. As the ego vehicle closes the longitudinal distance to slower traffic ($t=60\sim130$), the short-term memory module is progressively activated (orange curve) to inject multimodal priors, dynamically diversifying the search space to prevent policy collapse. Crucially, during the core lane-change execution ($t>130$), the planner encounters highly coupled multi-agent interactions. Here, epistemic uncertainty naturally spikes due to unpredictable adjacent behaviors. Rather than relying on a heuristic threshold, the trained gating network autonomously responds by shifting the allocation weights toward the rule-based fallback (green curve), effectively substituting neural exploration with deterministic kinematic constraints. Collectively, these dynamics validate the hierarchical defense architecture: blending implicit strategies for efficiency in predictable domains, while autonomously enforcing an explicit kinematic safety shield under elevated cognitive uncertainty.


\paragraph{Multimodal Preservation via GMM-CCEM}
Figure \ref{WM_horizon} depicts the temporal evolution of the action space during an overtaking maneuver across discrete time steps $t \in \{1, 15, 30, 45\}$. At each specific step, the planner executes a single optimization cycle, yielding numerous parallel-sampled $H$-horizon trajectory candidates. The bottom row visualizes the initial-step actions of these sequences within the steering-acceleration plane. During the early interaction phase ($t \in \{1, 15\}$), proximity to a slower preceding vehicle induces a pronounced topological bifurcation, splitting the feasible action space into distinct semantic modes: a ``decelerating lane-keeping'' cluster and an ``accelerating lane-changing'' cluster. Constrained by a unimodal assumption, the standard CEM mean (indicated by the green `X') erroneously converges to the arithmetic average of these disjoint modes, thereby collapsing into a kinematically hazardous region (e.g., accelerating without adequate steering).

Conversely, the GMM-CCEM framework effectively preserves this bimodality. Geometrically, these distinct Gaussian modes are bounded by covariance confidence ellipses, whose principal orientations and semi-axis lengths are mathematically derived from the eigenvectors and the scaled square roots of the corresponding eigenvalues of the component covariance matrix $\Sigma_m$. As the temporal sequence advances, GMM-CCEM not only maintains clear mode separation but also facilitates the continuous tracking of the optimal maneuver center (indicated by the purple shift vectors). By $t \in \{30, 45\}$, as the lane change stabilizes, the action distribution dynamically adapts to the requisite steering and acceleration profiles. Ultimately, while macroscopic metrics evaluate overall task success, this instantaneous microscopic analysis explicitly elucidates the internal mechanism mitigating mode collapse, ensuring spatiotemporal consistency and strict physical viability throughout complex interactions.

\begin{figure*}[ht]
	\centering
	\includegraphics[width=1\textwidth]{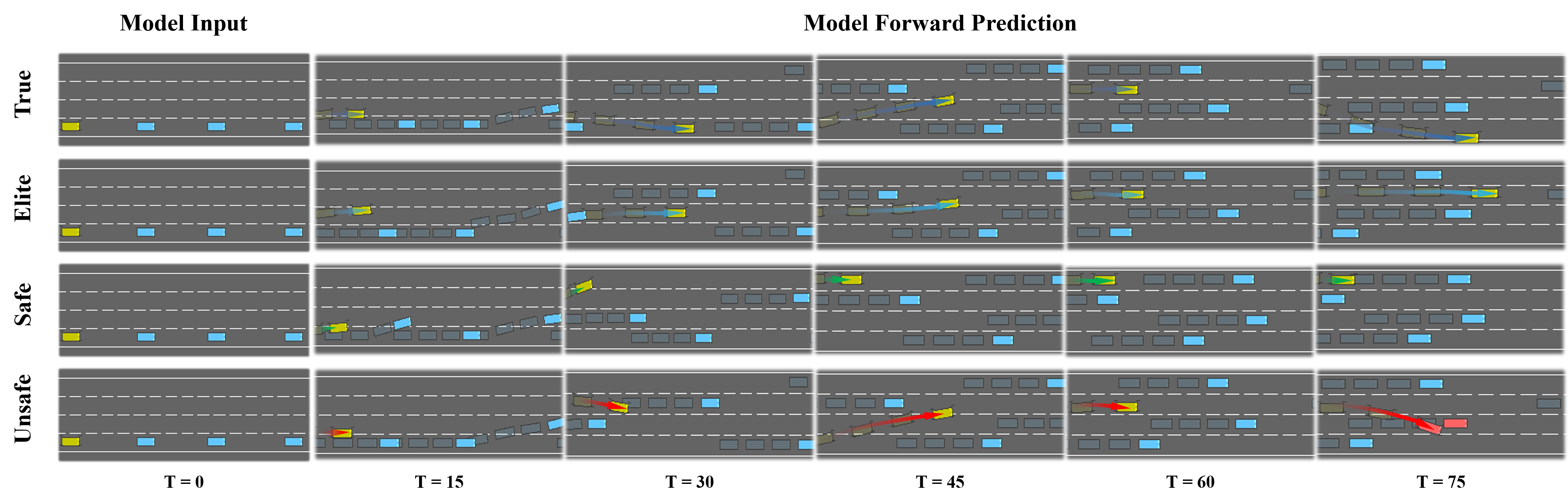}
	\caption{Visualization of receding-horizon latent rollouts within the learned world model.
		Rows illustrate the ground-truth environmental evolution (True), the optimized trajectory selected by the planner (Elite), a compliant candidate verifying safety constraints (Safe), and a counterfactual collision-prone rollout (Unsafe) marked in red, all evaluated across successive decision instants ($T=0$ to $75$).} 
	\label{WM_muti}
	\vspace{-15pt}
\end{figure*}

\paragraph{Verification of Receding-Horizon Latent Dynamics}
Figure \ref{WM_muti} details the verification of the world model's receding-horizon rollouts, providing a comparative analysis between the learned latent dynamics and the ground-truth environmental evolution. While the forward-predicted latent states inherently lack explicit physical parameters, their dynamic fidelity is rigorously validated by mapping the latent belief states back to the physical observation space via the trained observation decoder. Through the juxtaposition of the model-optimized trajectory (`Elite') and the actual scenario progression (`True') across extended time steps ($T=0 \sim 75$), the results substantiate that the world model maintains strict temporal consistency within the rollout window. Specifically, the model accurately anticipates the complex spatiotemporal interactions of surrounding vehicles alongside the ego vehicle's kinematic responses. Furthermore, the reliability of these $n$-step predictions is statistically fortified by the massive parallel rollout mechanism; the planner evaluates thousands of hypothetical trajectories and filters them based on expected costs. Crucially, the visualization of the `Unsafe' trajectory reveals the model's capacity for counterfactual risk reasoning. By correctly predicting the consequences of a hypothetical aggressive maneuver (e.g., a lateral collision, highlighted in red), the world model enables the preemptive pruning of dangerous action sequences purely within the latent space. Consequently, this imagination-driven mechanism ensures that the finalized trajectory statistically converges to a safe and physically viable solution.

\subsection{Summary of Experimental Analysis}
Collectively, the comprehensive microscopic and macroscopic evaluations validate the core mechanisms of the proposed WM-RMoE framework. Specifically, the microscopic visualizations serve distinct and critical analytical purposes: the temporal analysis of expert allocation substantiates the gating network's learned capacity to autonomously shift toward a rule-based safety shield under highly coupled interactions, thereby eliminating the reliance on heuristic uncertainty thresholds. Concurrently, the visualization of receding-horizon rollouts explicitly demonstrates the world model's capability for counterfactual risk reasoning, verifying that the latent dynamic predictions maintain high physical fidelity for forward planning. By tightly integrating this predictive foresight with GMM-CCEM's multimodal trajectory refinement, the system seamlessly unifies proactive risk anticipation with reactive kinematic constraints. This rigorous architectural integration effectively addresses the challenges of intricate driving scenarios, ensuring strict kinematic safety, adaptability, and operational efficiency in complex interactive traffic environments.

\section{CONCLUSION}
This paper presented the World Model Risk-aware Mixture-of-Experts framework, a novel architecture designed to reconcile the trade-off between driving efficiency and safety in stochastic highway environments. By synergizing latent space foresight with a heterogeneous expert policy, the proposed method facilitates a paradigm shift from myopic single-step constraint satisfaction to farsighted trajectory-level risk mitigation. Quantitative experiments demonstrate that WM-RMoE significantly outperforms state-of-the-art baselines across varying traffic densities, exhibiting superior collision avoidance capabilities and decision robustness. Crucially, the integration of the GMM-CCEM trajectory optimizer effectively alleviates the mode collapse issue inherent in uni-modal planners, preserving multimodal solution spaces during complex interactions. Furthermore, the dynamic gating mechanism ensures adaptive switching between rule-based safety priors and learning-based efficiency optimization, thereby enhancing the system's plasticity. 

Future work will extend this architecture by exploring centralized parallelized rollouts to enhance computational fidelity, integrating personalized driver profiles to capture diverse behavioral semantics, and deploying the system on physical autonomous platforms to validate real-world kinematic consistency.

\bibliographystyle{IEEEtran}
\bibliography{references}

@article{Schwarting2018Planning,
	author    = {Wilko Schwarting and Javier Alonso-Mora and Daniela Rus},
	title     = {Planning and Decision-Making for Autonomous Vehicles},
	journal   = {Annu. Rev. Control Robot. Auton. Syst.},
	volume    = {1},
	pages     = {187--210},
	year      = {2018},
	publisher = {Annual Reviews}
}

@inproceedings{ShalevShwartz2016Safe,
	author    = {Shai Shalev-Shwartz and Shaked Shammah and Amnon Shashua},
	title     = {Safe, Multi-Agent, Reinforcement Learning for Autonomous Driving},
	booktitle = {Proc. IEEE Intell. Veh. Symp. (IV)},
	year      = {2016},
	pages     = {1--8},
	address   = {Gothenburg, Sweden}
}

@article{Paden2016Survey,
	author    = {Brian Paden and Michal {\v{C}}{\'{a}}p and Sze Zheng Yong and Dmitry Yershov and Emilio Frazzoli},
	title     = {A Survey of Motion Planning and Control Techniques for Self-Driving Urban Vehicles},
	journal   = {IEEE Trans. Intell. Veh.},
	volume    = {1},
	number    = {1},
	pages     = {33--55},
	year      = {2016},
	publisher = {IEEE}
}

@article{Lefevre2014Survey,
	author    = {St{\'e}phanie Lef{\`e}vre and Dizan Vasquez and Christian Laugier},
	title     = {A Survey on Motion Prediction and Risk Assessment for Intelligent Vehicles},
	journal   = {Robot. Auton. Syst.},
	volume    = {58},
	number    = {3},
	pages     = {219--239},
	year      = {2014},
	publisher = {Elsevier}
}

@inproceedings{Fujimoto2018TD3,
	author    = {Scott Fujimoto and Herke van Hoof and David Meger},
	title     = {Addressing Function Approximation Error in Actor-Critic Methods},
	booktitle = {Proc. Int. Conf. Mach. Learn. (ICML)},
	year      = {2018},
	pages     = {1587--1596}
}

@article{Schulman2017PPO,
	author    = {John Schulman and Filip Wolski and Prafulla Dhariwal and Alec Radford and Oleg Klimov},
	title     = {Proximal Policy Optimization Algorithms},
	journal   = {arXiv preprint arXiv:1707.06347},
	year      = {2017}
}

@inproceedings{Haarnoja2018SAC,
	author    = {Tuomas Haarnoja and Aurick Zhou and Pieter Abbeel and Sergey Levine},
	title     = {Soft Actor-Critic: Off-Policy Maximum Entropy Deep Reinforcement Learning with a Stochastic Actor},
	booktitle = {Proc. Int. Conf. Mach. Learn. (ICML)},
	year      = {2018},
	pages     = {1861--1870}
}

@inproceedings{Kendall2019LearningSafety,
	author    = {Alex Kendall and Yarin Gal},
	title     = {What Uncertainties Do We Need in {Bayesian} Deep Learning for Computer Vision?},
	booktitle = {Adv. Neural Inf. Process. Syst. (NeurIPS)},
	year      = {2017},
	pages     = {5574--5584}
}

@book{Altman1999CMDP,
	title={Constrained Markov Decision Processes},
	author={Altman, Eitan},
	volume={7},
	year={1999},
	publisher={CRC press}
}

@article{Ray2019SafeExploration,
	author  = {Alex Ray and Joshua Achiam and Dario Amodei},
	title   = {Benchmarking Safe Exploration in Deep Reinforcement Learning},
	journal = {arXiv preprint arXiv:1910.01708},
	year    = {2019}
}

@article{Garcia2015SafeRL,
	author    = {Javier Garc\'{\i}a and Fernando Fern\'{a}ndez},
	title     = {A Comprehensive Survey on Safe Reinforcement Learning},
	journal   = {J. Mach. Learn. Res.},
	volume    = {16},
	number    = {1},
	pages     = {1437--1480},
	year      = {2015}
}

@inproceedings{Achiam2017CPO,
	author    = {Joshua Achiam and David Held and Aviv Tamar and Pieter Abbeel},
	title     = {Constrained Policy Optimization},
	booktitle = {Proc. Int. Conf. Mach. Learn. (ICML)},
	year      = {2017},
	pages     = {22--31}
}

@inproceedings{Tessler2019RCPO,
	author    = {Chen Tessler and Daniel J. Mankowitz and Shie Mannor},
	title     = {Reward Constrained Policy Optimization},
	booktitle = {Proc. Int. Conf. Learn. Representations (ICLR)},
	year      = {2019}
}

@inproceedings{Stooke2020PID,
	author    = {Adam Stooke and Joshua Achiam and Pieter Abbeel},
	title     = {Responsive Safety in Reinforcement Learning by {PID} Lagrangian Methods},
	booktitle = {Proc. Int. Conf. Mach. Learn. (ICML)},
	year      = {2020},
	pages     = {9133--9143}
}

@inproceedings{Chow2018Lyapunov,
	author    = {Yinlam Chow and Ofir Nachum and Edgar Duenez-Guzman and Mohammad Ghavamzadeh},
	title     = {A {Lyapunov}-based Approach to Safe Reinforcement Learning},
	booktitle = {Adv. Neural Inf. Process. Syst. (NeurIPS)},
	year      = {2018}
}

@inproceedings{Hafner2019PlaNet,
	author    = {Danijar Hafner and Timothy Lillicrap and Ian Fischer and Ruben Villegas and David Ha and Honglak Lee and James Davidson},
	title     = {Learning Latent Dynamics for Planning from Pixels},
	booktitle = {Proc. Int. Conf. Mach. Learn. (ICML)},
	year      = {2019},
	pages     = {2555--2565}
}

@inproceedings{Hafner2020Dreamer,
	author    = {Danijar Hafner and Timothy Lillicrap and Jimmy Ba and Mohammad Norouzi},
	title     = {Dream to Control: Learning Behaviors by Latent Imagination},
	booktitle = {Proc. Int. Conf. Learn. Representations (ICLR)},
	year      = {2020}
}

@inproceedings{Hafner2021DreamerV2,
	author    = {Danijar Hafner and Timothy Lillicrap and Mohammad Norouzi and Jimmy Ba},
	title     = {Mastering Atari with Discrete World Models},
	booktitle = {Proc. Int. Conf. Learn. Representations (ICLR)},
	year      = {2021}
}

@article{Hafner2023DreamerV3,
	author    = {Danijar Hafner and Jurgis Pasukonis and Jimmy Ba and Timothy Lillicrap},
	title     = {Mastering Diverse Domains through World Models},
	journal   = {arXiv preprint arXiv:2301.04104},
	year      = {2023}
}

@article{Zhang2021LVM,
	author    = {Yi Zhang and Yao Mu and Yi Yang and Yang Guan and Shengbo Eben Li and Qi Sun and Jianyu Chen},
	title     = {Steadily Learn to Drive with Virtual Memory},
	journal   = {arXiv preprint arXiv:2102.08072},
	year      = {2021}
}

@article{Gao2024SEM2,
	author    = {Zizhang Gao and Yao Mu and Cheng Chen and Jingyuan Duan and Ping Luo and Yang Lu and Shengbo Eben Li},
	title     = {Enhance Sample Efficiency and Robustness of End-to-End Urban Autonomous Driving via Semantic Masked World Model},
	journal   = {IEEE Trans. Intell. Transp. Syst.},
	volume    = {25},
	number    = {6},
	pages     = {5678--5692},
	year      = {2024},
	publisher = {IEEE}
}

@inproceedings{Pan2022IsoDream,
	author    = {Mowen Pan and Xiaogang Zhu and Yu Wang and Xiaokang Yang},
	title     = {Iso-Dream: Isolating and Leveraging Noncontrollable Visual Dynamics in World Models},
	booktitle = {Adv. Neural Inf. Process. Syst. (NeurIPS)},
	year      = {2022},
	volume    = {35},
	pages     = {25298--25310}
}

@inproceedings{Huang2024SafeDreamer,
	author    = {Weidong Huang and Jiaming Ji and Chunhe Xia and Borui Zhang and Yaodong Yang},
	title     = {SafeDreamer: Safe Reinforcement Learning with World Models},
	booktitle = {Proc. Int. Conf. Learn. Representations (ICLR)},
	year      = {2024}
}

@inproceedings{Janner2019MBPO,
	author    = {Michael Janner and Justin Fu and Marvin Zhang and Sergey Levine},
	title     = {When to Trust Your Model: Model-Based Policy Optimization},
	booktitle = {Adv. Neural Inf. Process. Syst. (NeurIPS)},
	year      = {2019},
	pages     = {12519--12530}
}

@inproceedings{Liu2025ITSC,
	author    = {Yongzhi Liu and Sunan Zhang and Changfeng Shen and Xiangwei Zhang and Weichao Zhuang},
	title     = {Risk-Aware Dual-Policy Coordination with World Model for Safe and Adaptive Highway Autonomous Driving},
	booktitle = {Proc. IEEE Intell. Transp. Syst. Conf. (ITSC)},
	year      = {2025},
	address   = {Gold Coast, Australia},
	month     = {Nov.}
}

@article{Zhang2026MeUAL,
	author    = {Sunan Zhang and Bo Li and Biao Chen and B. Hu and Cheng Sun and Weichao Zhuang},
	title     = {MeUAL: Model-Enhanced Uncertainty-Aware Safe Reinforcement Learning for Safety-Critical Autonomous Highway Overtaking},
	journal   = {IEEE Trans. Intell. Transp. Syst.},
	year      = {2026},
	note      = {to be published}
}

@article{Zhang2024Integration,
	author    = {Sunan Zhang and Weichao Zhuang and Bo Li and others},
	title     = {Integration of Planning and Deep Reinforcement Learning in Speed and Lane Change Decision-Making for Highway Autonomous Driving},
	journal   = {IEEE Trans. Transp. Electrific.},
	volume    = {11},
	number    = {1},
	pages     = {521--535},
	year      = {2024},
	month     = {Feb.}
}

@inproceedings{Chai2019MultiPath,
	author    = {Yuning Chai and Benjamin Sapp and Mayank Bansal and Dragomir Anguelov},
	title     = {MultiPath: Multiple Probabilistic Anchor Trajectory Hypotheses for Behavior Prediction},
	booktitle = {Proc. Conf. Robot. Learn. (CoRL)},
	year      = {2019},
	pages     = {86--99}
}

@inproceedings{Codevilla2018CIL,
	author    = {Felipe Codevilla and Matthias M\"{u}ller and Alexey Dosovitskiy and Antonio L\'{o}pez and Vladlen Koltun},
	title     = {End-to-End Driving via Conditional Imitation Learning},
	booktitle = {Proc. IEEE Int. Conf. Robot. Autom. (ICRA)},
	year      = {2018},
	pages     = {4693--4700}
}

@inproceedings{Lepikhin2020GShard,
	author    = {Dmitry Lepikhin and HyoukJoong Lee and Yuanzhong Xu and Dehao Chen and Orhan Firat and Yanping Huang and Maxim Krikun and Noam Shazeer and Zhifeng Chen},
	title     = {GShard: Scaling Giant Models with Conditional Computation and Automatic Sharding},
	booktitle = {Proc. Int. Conf. Learn. Representations (ICLR)},
	year      = {2021}
}

@article{Fedus2021Switch,
	author    = {William Fedus and Barret Zoph and Noam Shazeer},
	title     = {Switch Transformers: Scaling to Trillion Parameter Models with Simple and Efficient Sparsity},
	journal   = {J. Mach. Learn. Res.},
	volume    = {23},
	number    = {120},
	pages     = {1--39},
	year      = {2022}
}

@article{Wan2025GEMINUS,
	author    = {Chuan Wan and Yasheng Cui and Jun Du and Shuang Yang and Yu Bai and Ping Yi and Na Li and Yuchao Huang},
	title     = {GEMINUS: Dual-aware Global and Scene-Adaptive Mixture-of-Experts for End-to-End Autonomous Driving},
	journal   = {arXiv preprint arXiv:2507.14456},
	year      = {2025}
}

@article{Feng2025ARTEMIS,
	author    = {Ruizheng Feng and Nuo Xi and Dongsheng Chu and Renzhi Wang and Zhaojian Deng and Ao Wang and Long Lu and Jun Wang and Yuchao Huang},
	title     = {ARTEMIS: Autoregressive End-to-End Trajectory Planning with Mixture of Experts for Autonomous Driving},
	journal   = {arXiv preprint arXiv:2504.19580},
	year      = {2025}
}

@article{Cui2025MoPE,
	author    = {Yasheng Cui and Shuang Yang and Chuan Wan and Xinyu Li and Jin Xing and Yan Zhang and Yuchao Huang and Hong Chen},
	title     = {Continual Adaptation for Autonomous Driving with the Mixture of Progressive Experts Network},
	journal   = {arXiv preprint arXiv:2502.05943},
	year      = {2025}
}

@article{Xu2025KDP,
	author    = {Chenghao Xu and Jian Liu and Yifan Guo and Peng Hang and Jian Sun},
	title     = {A Knowledge-Driven Diffusion Policy for End-to-End Autonomous Driving Based on Expert Routing},
	journal   = {arXiv preprint arXiv:2509.04853},
	year      = {2025}
}

@inproceedings{Kingma2014VAE,
	title={Auto-Encoding Variational Bayes},
	author={Kingma, Diederik P and Welling, Max},
	booktitle={2nd International Conference on Learning Representations, {ICLR} 2014},
	year={2014},
	address={Banff, AB, Canada},
	month={April}
}

@book{Sutton2018Planning,
	author    = {Richard S. Sutton and Andrew G. Barto},
	title     = {Reinforcement Learning: An Introduction},
	edition   = {2nd},
	publisher = {MIT Press},
	address   = {Cambridge, MA, USA},
	year      = {2018}
}

@book{Buehler2009Planning,
	editor    = {Martin Buehler and Karl Iagnemma and Sanjiv Singh},
	title     = {The DARPA Urban Challenge: Autonomous Vehicles in City Traffic},
	series    = {Springer Tracts in Advanced Robotics},
	volume    = {56},
	publisher = {Springer},
	address   = {Berlin, Heidelberg},
	year      = {2009}
}

@inproceedings{Williams2017MPPI,
	author    = {Grady Williams and Andrew Aldrich and Evangelos A. Theodorou},
	title     = {Model Predictive Path Integral Control: From Theory to Parallel Computation},
	booktitle = {Proc. IEEE Int. Conf. Robot. Autom. (ICRA)},
	year      = {2017},
	pages     = {663--670}
}

@article{Mayne2000MPC,
	author    = {David Q. Mayne and James B. Rawlings and Christopher V. Rao and Pierre O. M. Scokaert},
	title     = {Constrained Model Predictive Control: Stability and Optimality},
	journal   = {Automatica},
	volume    = {36},
	number    = {6},
	pages     = {789--814},
	year      = {2000}
}

@article{Treiber2000IDM,
	author    = {Martin Treiber and Ansgar Hennecke and Dirk Helbing},
	title     = {Congested Traffic States in Empirical Observations and Microscopic Simulations},
	journal   = {Phys. Rev. E},
	volume    = {62},
	number    = {2},
	pages     = {1805--1824},
	year      = {2000},
	publisher = {American Physical Society}
}

@article{Kesting2007MOBIL,
	author    = {Arne Kesting and Martin Treiber and Dirk Helbing},
	title     = {General Lane-Changing Model {MOBIL} for Car-Following Models},
	journal   = {Transp. Res. Rec.},
	volume    = {1999},
	number    = {1},
	pages     = {86--94},
	year      = {2007},
	publisher = {SAGE Publications}
}

@misc{Leurent2018Highway,
	author       = {Edouard Leurent},
	title        = {An Environment for Autonomous Driving Decision-Making},
	year         = {2018},
	publisher    = {GitHub},
	journal      = {GitHub repository},
	howpublished = {\url{https://github.com/eleurent/highway-env}},
	note         = {Accessed: 2024-05-20} 
}

@article{Li2022Continuous,
	title={Continuous Decision-Making in Lane Changing and Overtaking Maneuvers for Unmanned Vehicles: A Risk-Aware Reinforcement Learning Approach with Task Decomposition},
	author={Li, Bai and Ouyang, Yankai and Li, Li and Zhang, Youmin},
	journal={IEEE Transactions on Intelligent Vehicles},
	volume={7},
	number={3},
	pages={558--571},
	year={2022},
	publisher={IEEE},
	doi={10.1109/TIV.2022.3169183}
}

\end{document}